\UseRawInputEncoding
\documentclass[11pt]{article}
\usepackage[top=1in, bottom=1in, left=1in, right=1in]{geometry}

\newcommand{\agentname}{\texttt{CudaForge}}
    \PassOptionsToPackage{numbers, compress,sort}{natbib}

\usepackage[most]{tcolorbox} 
\usepackage{listings}
\usepackage[T1]{fontenc}
\usepackage{inconsolata}

\lstdefinestyle{nice}{
  basicstyle=\ttfamily\small,
  numbers=left,
  numberstyle=\scriptsize\color{blue},
  numbersep=8pt,
  showstringspaces=false,
  breaklines=true,
  frame=single,
  columns=fullflexible,
  tabsize=2,
  keepspaces=true,
  upquote=true,
  keywordstyle=\bfseries,
  commentstyle=\itshape,
}

\tcbset{
  mybox/.style={
    enhanced,
    breakable,
    colback=blue!4,             
    colframe=blue!65!black,     
    colbacktitle=blue!70!black,
    coltitle=white,            
    fonttitle=\bfseries,        
    boxrule=0.8pt,
    arc=2mm,
    left=2mm, right=2mm, top=1mm, bottom=1mm,
    title style={font=\bfseries},
  },
    redbox/.style={
    enhanced,
    breakable,
    colback=red!5,
    colframe=red!70!black,
    colbacktitle=red!75!black,
    coltitle=white,
    fonttitle=\bfseries,
    boxrule=0.8pt,
    arc=2mm,
    left=2mm, right=2mm, top=1mm, bottom=1mm,
    title style={font=\bfseries},
  }
}

\newcommand{\Metric}[1]{\texttt{\seqsplit{#1}}}
\usepackage{pgfplots}
\usepackage{float}
\usepackage{hyperref}
\usepackage{amssymb}
\usepackage{siunitx}
\usepackage{listings}
\usepackage{xcolor}

\usepackage{wrapfig}   

\usepackage{booktabs,longtable, array} 
\usepackage{siunitx}                  
\usepackage{seqsplit}                 
\usepackage{pdflscape}   
\usepackage[utf8]{inputenc}

\usepackage{tikz}
\usepackage{pgfplots}
\pgfplotsset{compat=1.18}
\usepackage{multirow}
\usepackage{graphicx}
\usepackage{natbib}[numbers,sort,compress]
\usepackage[ruled]{algorithm2e}

\usepackage[utf8]{inputenc} 
\usepackage[T1]{fontenc}    
\usepackage{hyperref}       
\usepackage{url}            
\usepackage{booktabs}       
\usepackage{amsfonts}       
\usepackage{nicefrac}       
\usepackage{microtype}      
\usepackage{xcolor}         
\usepackage{subcaption}

\usepackage{wrapfig}
\usepackage{multirow,mathtools}
\usepackage{pifont}
\usepackage{color, colortbl}

\usepackage{blindtext}
\usepackage{lipsum}

\usepackage{multirow}
\usepackage{makecell}
\usepackage{graphicx}
\usepackage{listings}

\usepackage{bbm}

\usepackage [english]{babel}
\usepackage [autostyle, english = american]{csquotes}
\usepackage[nottoc]{tocbibind}

\usepackage{pifont}
\usepackage{url}
\usepackage[most]{tcolorbox}

\usepackage{lipsum}
\usepackage{xcolor}
\usepackage{wrapfig}
\usepackage{booktabs}

\usepackage{bbold}

\usepackage[most]{tcolorbox}
\usepackage{pifont}
\usepackage{multirow}
\usepackage{booktabs}
\usepackage{colortbl}

\definecolor{Gray}{gray}{0.93}
\definecolor{Orange}{rgb}{1,0.5,0}
\definecolor{DGray}{gray}{0.83}
\definecolor{modelrowcolor}{RGB}{204,229,255}
\definecolor{darkergreen}{RGB}{1, 50, 32}

\usepackage{color, colortbl}

\usepackage{pifont}

\usepackage{color, colortbl}
\definecolor{Gray}{gray}{0.93}
\definecolor{Orange}{rgb}{1,0.5,0}
\definecolor{Green}{rgb}{0,0.80,0}
\definecolor{Blue}{rgb}{0,0,0.92}
\definecolor{Red}{rgb}{0.90,0,0}
\definecolor{DGray}{gray}{0.83}
\definecolor{LightCyan}{rgb}{0.88,1,1}

\definecolor{bluegray}{rgb}{0.4, 0.6, 0.8}
\definecolor{ceruleanblue}{rgb}{0.16, 0.32, 0.75}

\usepackage{amsmath,amsfonts,bm}

\def\eqref#1{(\ref{#1})}

\def\1{\bm{1}}

\DeclareMathAlphabet{\mathsfit}{\encodingdefault}{\sfdefault}{m}{sl}
\SetMathAlphabet{\mathsfit}{bold}{\encodingdefault}{\sfdefault}{bx}{n}

\everydisplay{\small}

\title{
CudaForge: An Agent Framework with Hardware Feedback for CUDA Kernel Optimization
}

\author{%
  \large
  Zijian Zhang\thanks{Co-first author. Equal contribution; order decided by a coin toss.}\quad
  Rong Wang\footnotemark[1]\quad
  Shiyang Li\quad
  Yuebo Luo\\[0.3cm]
  Mingyi Hong\thanks{Corresponding author.}\quad
  Caiwen Ding\footnotemark[2]\\[0.8em]
  \normalsize University of Minnesota, Twin Cities\\[0.3em]
  \texttt{\{zha00175, wan00559, li004074, luo00466, mhong, dingc\}@umn.edu}%
}

\usetikzlibrary{shapes.geometric}

\date{}
\begin{document}
\maketitle

\begin{abstract}
Developing efficient CUDA kernels is increasingly critical for AI applications such as large-scale LLM training. However, manual kernel design is both costly and time-consuming, motivating automatic approaches that leverage LLMs for code generation. Existing methods for automatic kernel generation, however, often produce low-efficiency kernels, incur high computational overhead, and fail to generalize across different settings. 

In this work, we propose \agentname, a training-free multi-agent workflow for CUDA kernel generation and optimization. Our workflow is inspired by the iterative workflow of human experts, which contains steps such as developing initial kernels, testing correctness, analyzing hardware feedback, and iterative improvement.
More specifically, \agentname~employs two LLM agents: a Coder and a Judge, that iteratively generate, correct, and optimize CUDA kernels, while integrating hardware feedback such as Nsight Compute (NCU) metrics. In our extensive evaluations, we show that \agentname, by leveraging base models like OpenAI-o3, achieves 97.6\% correctness of generated kernels and an average 1.68$\times$ speedup over PyTorch baselines, substantially surpassing state-of-the-art models including OpenAI-o3 and Kevin on KernelBench, while further scaling up maximum iteration rounds increases \agentname's performance to 2.27$\times$ speedup, showing its strong capability in practice. Beyond accuracy and speed, \agentname~demonstrates strong generalization across GPUs (A100, RTX 6000, 4090, 3090) and base models (OpenAI-o3, GPT-5, gpt-oss-120B, Claude-Sonnet-4, QwQ-32B), while maintaining high efficiency. In particular, generating an optimized kernel takes about $26.5$ minutes on one RTX6000 and incurs about \$ 0.3 API cost, which is significantly cheaper than existing agentic work that costs 6 H100 hours and \$ 5 API cost per kernel.  Our results highlight that multi-agent, training-free workflows can enable cost-effective, generalizable, and high-performance CUDA kernel optimization. Code available at \url{https://github.com/OptimAI-Lab/CudaForge}
\end{abstract}

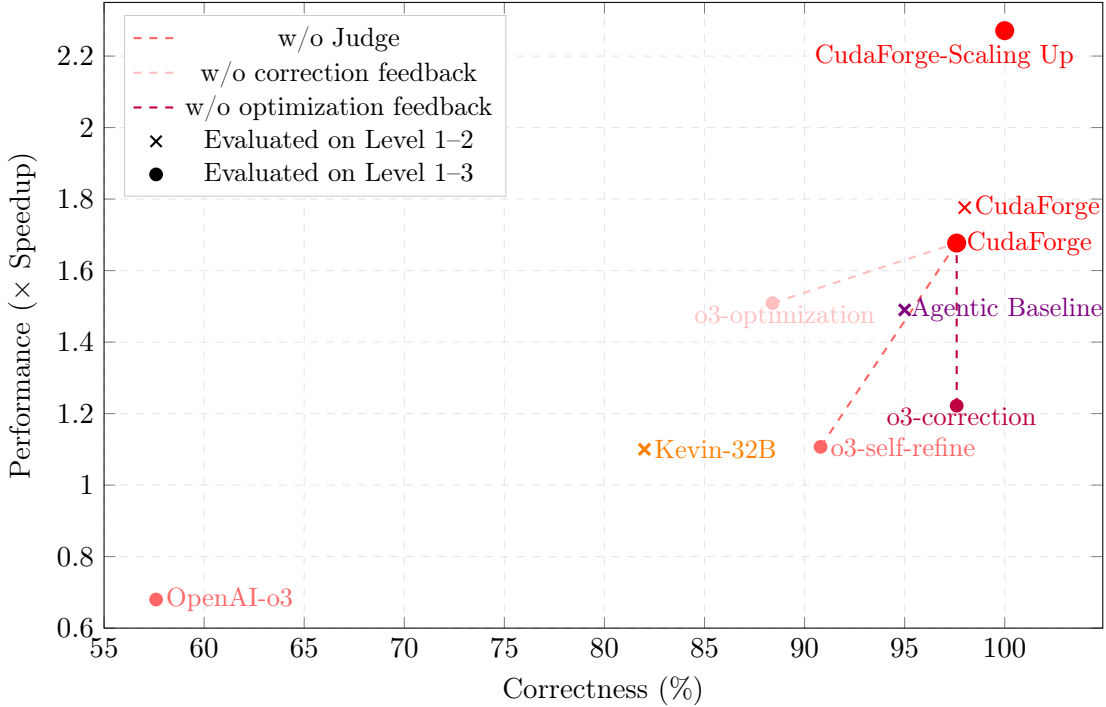
\begin{figure}[t]
\centering
\begin{tikzpicture}
\begin{axis}[
  width=0.9\linewidth,
  height=0.6\linewidth,
  xlabel={Correctness (\%)},
  ylabel={Performance (× Speedup)},
  xmin=55, xmax=104.9,
  ymin=0.6, ymax=2.35,
  grid=both,
  grid style={dashed,gray!20},
  legend style={at={(0.02,0.98)},anchor=north west,font=\small,fill=white,draw=gray!40},
  every axis plot/.append style={thick}
]

\addplot[red!60!white,mark=none,dashed] coordinates {
  (90.8,1.107)
  (97.6,1.677)
};
\addlegendentry{w/o Judge}

\addplot[pink,mark=none,dashed] coordinates {
  (88.4,1.509)
  (97.6,1.677)
};
\addlegendentry{w/o correction feedback}

\addplot[purple,mark=none,dashed] coordinates {
  (97.6,1.222)
  (97.6,1.677)
};
\addlegendentry{w/o optimization feedback}
\addlegendimage{only marks,mark=x,mark size=3pt,black}
\addlegendentry{Evaluated on Level 1–2}

\addlegendimage{only marks,mark=*,mark size=2.2pt,black}
\addlegendentry{Evaluated on Level 1–3}

\addplot[only marks,mark=x,mark size=3pt,very thick,orange] coordinates {(82.0,1.10)};
\addplot[only marks,mark=x,mark size=3pt,very thick,violet] coordinates {(95.0,1.490)};

\addplot[only marks,mark=*,mark size=2.2pt,red!60!white] coordinates {(57.6,0.680)};
\addplot[only marks,mark=*,mark size=2.2pt,red!60!white] coordinates {(90.8,1.107)};

\addplot[only marks,mark=*,mark size=2.2pt,purple] coordinates {(97.6,1.222)};

\addplot[only marks,mark=*,mark size=2.2pt,pink] coordinates {(88.4,1.509)};

\addplot[only marks,mark=*,mark size=3.2pt,red] coordinates {(97.6,1.677)};

\addplot[only marks,mark=x,mark size=3.2pt,red] coordinates {(98,1.776)};

\addplot[only marks,mark=*,mark size=3.2pt,red] coordinates {(100,2.271)};

\node[anchor=west,font=\small,red!60!white] at (axis cs:57.6,0.68) {OpenAI-o3};
\node[anchor=west,font=\small,red!60!white] at (axis cs:90.8,1.107) {o3-self-refine};
\node[anchor=west,font=\small,purple] at (axis cs:93.6,1.192) {o3-correction};
\node[anchor=west,font=\small,pink]  at (axis cs:84.0,1.472) {o3-optimization};
\node[anchor=west,font=\small, red] at (axis cs:97.6,1.677) {CudaForge};
\node[anchor=west,font=\small, red] at (axis cs:98.0,1.776) {CudaForge};
\node[anchor=west,font=\small, red] at (axis cs:90.0,2.200) {CudaForge-Scaling Up};
\node[anchor=west,font=\small,orange] at (axis cs:82.0,1.10) {Kevin-32B};
\node[anchor=west,font=\small,violet] at (axis cs:94.8,1.490) {Agentic Baseline};

\end{axis}
\end{tikzpicture}
\caption{\agentname~achieves state-of-the-art results on KernelBench in both correctness and performance, 
surpassing RL-based methods such as Kevin-32B~\citep{kevin-multi-turn-rl}, 
the agentic baseline~\citep{lange2025robustagenticcudakernel}, 
and OpenAI-o3~\citep{openai2025o3systemcard}. 
To further evaluate the effectiveness of our design, 
we additionally develop three customized variants of OpenAI-o3: {o3-self-refine}, {o3-correction}, and {o3-optimization}, which serve as baselines for ablation comparison. Scaling up maximum iteration rounds(\agentname-Scaling Up) further improves \agentname's performance to 2.27$\times$ speedup.
Experimental details are provided in Section~\ref{exp}.}
\label{fig:evolution_path}
\end{figure}

\vspace{-14pt}

\section{Introduction}

\noindent{\bf Motivation.}
CUDA has become the \textit{de facto} standard for deep learning training because modern frameworks such as PyTorch and TensorFlow are deeply integrated with NVIDIA’s optimized GPU libraries~\citep{nvidia2025cudnn}. 
Efficient CUDA kernels are crucial for accelerating deep learning workloads~\citep{dao2022flashattention,dao2023flashattention2} .

However, developing high-efficiency CUDA kernels is known to be challenging with a steep learning curve, requiring deep expertise in GPU architectures and parallel programming \citep{onegraph}.
For example, it took more than 2 years from the debut of the Hopper GPU architecture to the release of FlashAttentionV3~\citep{shah2024flashattention3fastaccurateattention}, which is specially designed for Hopper GPUs. 

This high development barrier has driven growing interest in finding automated ways of generating highly efficient and customized CUDA kernels. For example, some work~\citep{DBLP:conf/pldi/TilletKC19}~\citep{chen2019learningoptimizetensorprograms} employs auto-tuning and evolutionary search to automatically explore kernel implementation spaces and optimize low-level parameters for specific hardware. More recently, there has been a growing interest in leveraging large language models (LLMs) to perform such tasks. LLM is believed to hold great promise in generating efficient and high-quality kernels, due to its capability of code generation in other domains, such as Python and C++~\citep{dong2025surveycodegenerationllmbased,jiang2024surveylargelanguagemodels}.

\noindent{\bf{ Existing Works and Key Challenges.}} Generally, using LLMs for CUDA kernel generation is still in an early stage. In KernelBench~\citep{ouyang2025kernelbenchllmswriteefficient}, the authors attempt to directly use state-of-the-art (SOTA) models, such as OpenAI-o1 and Claude-3.5-Sonnet, to generate kernels. However, it has been observed that these SOTA models still struggle to produce correct or performant kernels out of the box, revealing fundamental limitations of existing LLMs in this domain. 

To address this gap, recent studies have explored two main paradigms. The first approach is based on reinforcement learning (RL)~\citep{schulman2017proximalpolicyoptimizationalgorithms,shao2024deepseekmathpushinglimitsmathematical}. CUDA-L1~\citep{deepreinforce2025cudal1} and Kevin~\citep{kevin-multi-turn-rl} adopt RL to enhance LLMs' ability to generate correct and performant CUDA code. 

The second approach is based on AI agents. In particular, in an independent and contemporaneous work \citep{lange2025robustagenticcudakernel}\footnote{published on arxiv Sept 16th, 2025}, researchers have explored agentic frameworks at inference time. 
Agents project PyTorch methods into CUDA kernel designs, then the CUDA kernels are further refined by sampling new kernels and verification filtering.
This design effectively improves correctness in CUDA kernel generation without the high cost of RL training.

Despite these advances, several key challenges remain:

\noindent \textbf{(C1) Limited kernel efficiency.}  
    While RL-based methods improve LLMs' ability to generate CUDA kernels, their optimization capability remains insufficient.  
    For example, the kernels generated by Kevin-32B only achieve an average speedup of 1.10$\times$ over KernelBench Level 1-2, even after sampling 16 parallel trajectories with 8 refinement turns for each kernel  \citep{kevin-multi-turn-rl}.  
    As another example, CUDA-L1 often fails to directly optimize the CUDA kernels, but produces official implementations of PyTorch \citep{deepreinforce2025cudal1} (see Appendix~\ref{app:cuda-l1} for details).

\noindent \textbf{(C2) High training and inference cost.}  
    RL-based approaches such as \citep{deepreinforce2025cudal1,kevin-multi-turn-rl} require substantial computational resources and long training cycles, making them unsuitable for low-resource or rapid-prototyping settings.  
    In addition,  multi-stage agentic pipeline developed by \citep{lange2025robustagenticcudakernel} incurs high inference costs (about 6 H100 hours and \$5 API cost per kernel), which greatly limits its practical applicability of the approach.

\noindent \textbf{(C3) Lack of hardware feedback.}  
Human experts typically follow an iterative workflow to develop CUDA kernels through testing and refinement.  
They rely on hardware feedback like Nsight Compute (NCU)\footnote{Nsight Compute (NCU) is NVIDIA’s official kernel-level profiler for CUDA programs.} to identify bottlenecks and optimize kernels accordingly~\citep{10946796,cudaguide,w4a8gemm}.  
In contrast, RL-based approaches~\citep{deepreinforce2025cudal1,kevin-multi-turn-rl} train LLMs to directly generate or optimize kernels, but do not leverage hardware feedback at all.  
As a result, they rely on blind exploration during generation, lacking targeted guidance.  
This often leads to suboptimal kernel efficiency, limiting their practical applicability.

These challenges raise a natural question: \textit{Can we design a simple but effective hardware-aware approach that reliably produces efficient CUDA kernels at low cost?}

\begin{figure}[htbp]
  \centering
  \includegraphics[width=1.0\linewidth]{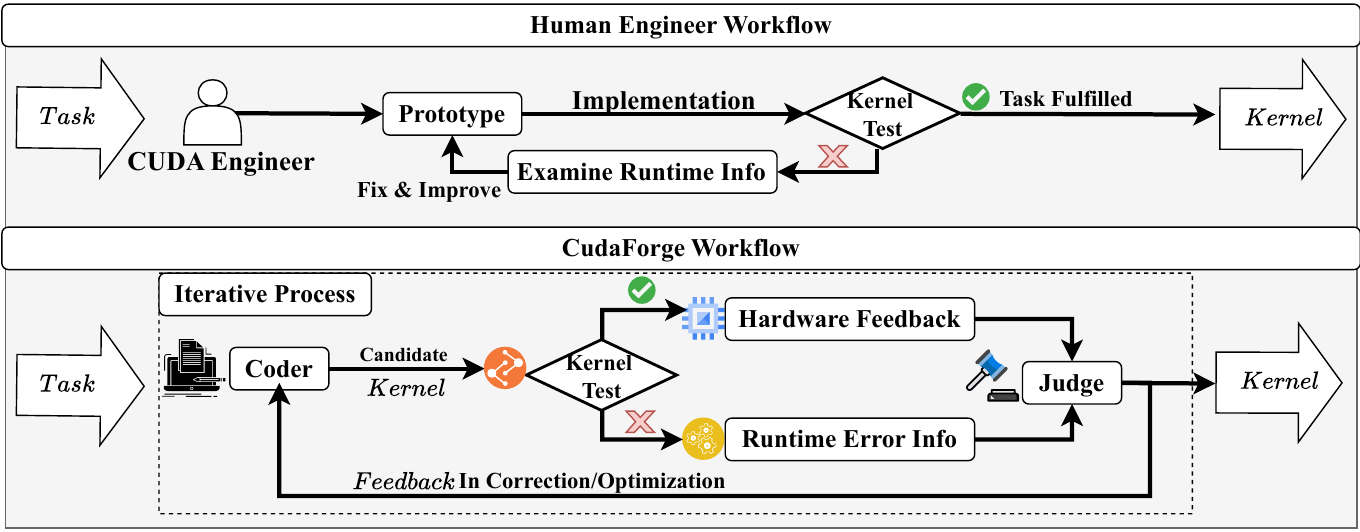}
  \caption{Comparison between human and \agentname~workflows.
Top: Human experts iteratively refine kernels by writing a prototype, testing it, and analyzing runtime feedback.
Bottom: \agentname~mimics human workflow with two specialized agents (Coder and Judge). The Coder generates candidate kernels, while the Judge analyzes runtime info and hardware feedback to provide correction or optimization feedback. The process iterates until it reaches maximum round $N$.}
  \label{fig:pipeline}
\end{figure}

\noindent{\bf Our Contributions.} 
To address these challenges, we propose \agentname, {\bf a simple,  effective, and low-cost} multi-agent workflow for CUDA kernel generation and optimization, as shown in Figure~\ref{fig:pipeline}. Our workflow is inspired by the iterative workflow of human experts~\citep{10946796,cudaguide,w4a8gemm}, which contains steps such as developing initial kernels, testing correctness, analyzing hardware feedback, and iterative improvement.

This workflow involves two specialized LLM agents that iteratively generate and optimize CUDA kernels: a Coder, which generates kernels given task instructions and Judge feedback, and a Judge, which analyzes kernels and hardware feedback to guide the Coder generation. One key novelty of \agentname~is its integration of external hardware feedback, including GPU specifications and Nsight Compute (NCU) metrics, enabling the Judge to identify performance bottlenecks like human experts and provide targeted optimization guidance to the Coder.

Compared to single-LLM approaches that  generate and evaluate code using the same LLM, our framework separates these roles into an {\it independent} Coder and Judge, enabling more specialized reasoning and more reliable iterative refinement.
Unlike RL-based methods, \agentname~is  training-free, avoiding the substantial cost of policy training. It is also  hardware-aware, allowing it to tailor CUDA kernel optimizations to the underlying system, making the proposed framework easily generalizable across different GPUs.
Finally, in contrast to existing multi-agent frameworks \citep{lange2025robustagenticcudakernel}, \agentname~is lightweight and cost-efficient, running in just 26.5 minutes on a single RTX 6000 GPU and about \$ 0.3 per kernel in API on average, while still achieving significantly better performance.

We evaluate \agentname~on 250 KernelBench tasks from Level 1 to Level 3. Though these tasks are challenging, \agentname~attains a 97.6\% correctness rate and delivers an average speedup of 1.68× over PyTorch baselines, which significantly outperforms advanced RL model like Kevin-32B, advanced frontier model like OpenAI-o3~\citep{openai2025o3systemcard} and Agentic Baseline~\citep{lange2025robustagenticcudakernel}, shown in Figure~\ref{fig:evolution_path}. Further, we have conducted comprehensive ablation studies of the features of \agentname, such as its effectiveness across multiple GPU architectures, its inference-time scalability by increasing the number of generation, and the effect of different base models. Overall, we observed that the proposed  \agentname~achieves robust performance in all these settings.

These findings highlight the key contribution of this work: The proposed LLM agent workflow \agentname~is simple but effective: at very low cost, it  develops performant CUDA kernels for many practical tasks, for a variety of GPU architectures and base models. It also exhibits strong test-time scaling capabilities where solution quality can improve substantially while increasing its iteration rounds. These results demonstrate \agentname's strong practical applicability.

\section{The CudaForge Framework for CUDA Kernel Optimization}\label{sec:proposed:approaches}
\vspace{-8pt}
\subsection{CudaForge Framework}\label{sec:overview}
 \vspace{-6pt}
Given a CUDA kernel generation task, the objective is to generate a kernel that is functionally equivalent to its PyTorch reference while achieving the lowest possible execution latency.

Inspired by the iterative workflow of human experts~\citep{10946796,cudaguide,w4a8gemm}, we design \agentname~as an iterative multi-agent framework, illustrated in Figure~\ref{fig:pipeline}.  The framework involves two independent agents: a \textbf{Coder} and a \textbf{Judge}. 
The Coder generates candidate kernels based on the task description and feedback from the Judge, while the Judge evaluates each candidate using the kernel itself, hardware feedback, and runtime information.

\begin{figure}[htbp]
  \centering
  \includegraphics[width=\linewidth]{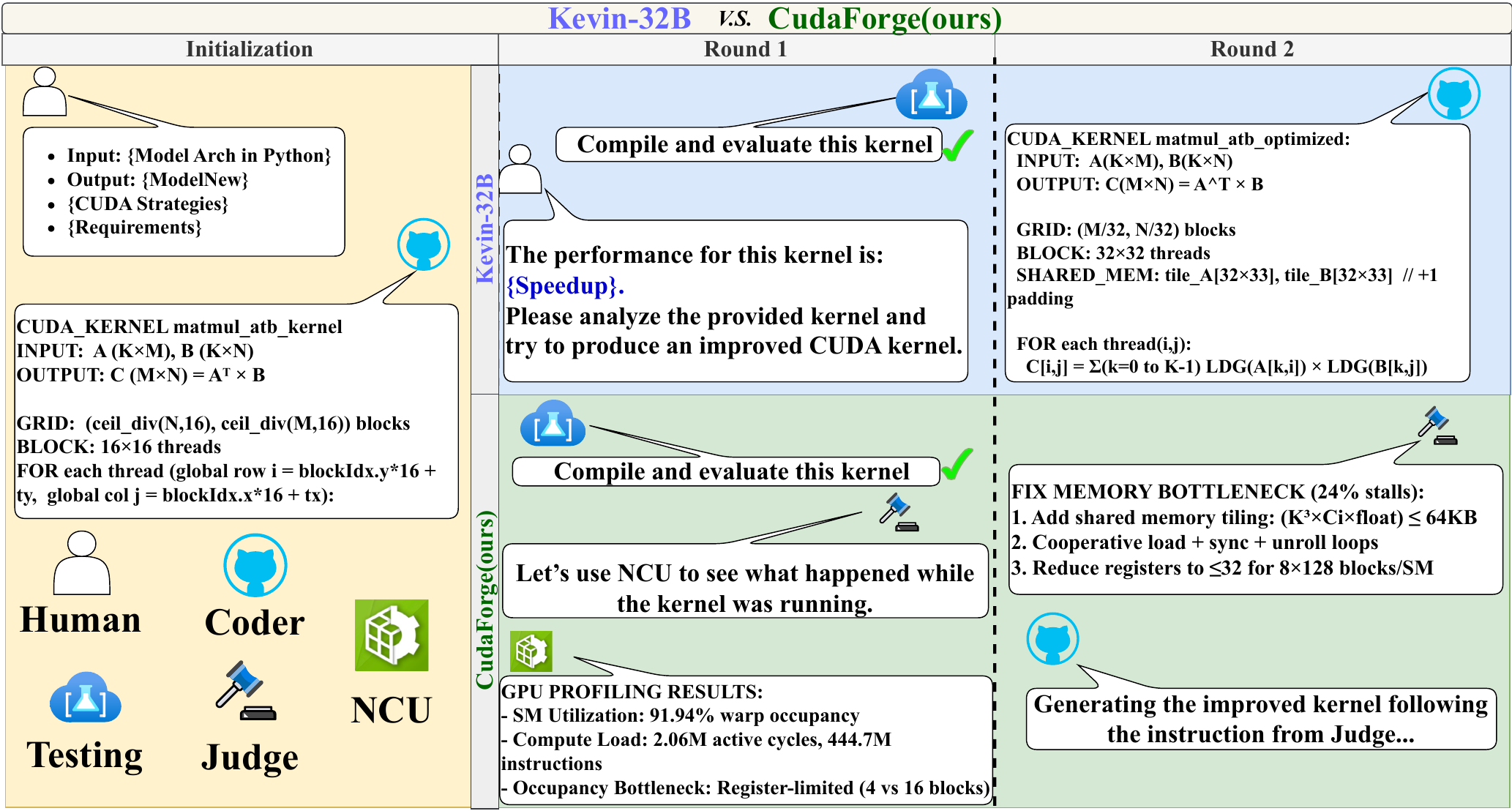}
  \caption{
The overview of how \agentname~optimizes kernels, compared with Kevin-32B. 
\textbf{Top:} the pipeline of the RL-based Kevin-32B, which relies solely on textual refinement and thus performs blind exploration. 
\textbf{Bottom:} our \agentname~workflow, which leverages hardware feedback to guide kernel optimization. 
When the Coder in \agentname~generates a correct candidate kernel in Round~1, the system profiles it using Nsight Compute (NCU) to obtain NCU metrics. 
In  Round~2, the Judge analyzes these metrics and GPU specifications to identify performance bottlenecks (e.g., register- or memory-limited) and provides targeted optimization feedback. 
The Coder then refines the kernel accordingly. 
Compared with Kevin-32B, which only refines based on speedup scores, our framework achieves more interpretable and effective performance improvements through hardware-aware iteration. 
}

  \label{fig:detail}
  \vspace{-8pt}
\end{figure}
\noindent 

More specifically, given a CUDA kernel generation task, the Coder first receives the task requirements and PyTorch reference implementation, then produces an initial candidate kernel.  
This kernel is compiled and executed on test cases to check its correctness.  
If it fails, the Judge inspects {\it runtime information} (e.g. compilation errors, mismatched outputs with the PyTorch reference) and analyzes the faulty kernel. It then returns {correction feedback} (e.g. missing header file) to guide the next round. Once a kernel candidate passes the  correctness test, the Judge profiles it with the NCU tool to obtain {NCU metrics} (e.g. memory throughput, occupancy, warp efficiency).  
Together with {GPU specifications}, these metrics form the {\it hardware feedback} that allows the Judge to identify the dominant bottleneck (e.g. compute-bound or memory-bound) and provide one specific {optimization feedback} (e.g. using shared memory) to the Coder.

In the next round, the Coder is prompted with the previous kernel, Judge feedback, and the original task requirements, and generates a corrected or optimized kernel.  
This process repeats for up to $N$ rounds, after which we select the most efficient correct kernel as the final solution.

\agentname~achieves reliability and efficiency through {\bf three key design choices}.
First, it adopts a two-agent system where the Coder focuses on generation and the Judge on evaluation, separating the “cognitive” load (See Section~\ref{Ablation-agents} for ablation study). The Coder receives only feedback from the Judge, while the Judge uses hardware and runtime information to guide generation and optimization. This division of labor mirrors human workflows and mitigates the risk of overlooking errors or inefficiencies.
Second, the framework follows an iterative optimization process, progressively correcting errors and improving efficiency across rounds. This enables stable refinement, especially on hard tasks.
Third, it explicitly incorporates hardware feedback, such as GPU specifications and NCU metrics, so that the Judge can pinpoint bottlenecks and provide actionable guidance to the Coder. This targeted optimization avoids blind exploration and ensures directed performance gains. 

\vspace{-10pt}
\subsection{Component Design}
\vspace{-4pt}
\paragraph{Design of Coder.} 
The {Coder} acts as the generative module in our framework, responsible for producing functionally correct and efficient CUDA kernel candidates given the task description, previous candidates, and feedback from the Judge. 
Directly generating a CUDA kernel from scratch is a highly challenging task for LLMs~\citep{ouyang2025kernelbenchllmswriteefficient}. 
Following the setting of KernelBench~\citep{ouyang2025kernelbenchllmswriteefficient}, we provide the Coder with a {one-shot demonstration} in the prompt, which includes an example pair of a PyTorch reference and its corresponding CUDA kernel implementation. 
This example serves as a structural and stylistic guide, helping the Coder understand the expected syntax, memory access patterns, and API usage required for valid kernel generation.

Moreover, we find that managing the Coder’s {memory scope} is crucial for stable and efficient generation. 
Prompting the model with the entire dialogue history introduces excessive context redundancy, often leading to hallucinated kernel code and higher API cost. 
To mitigate these issues, we adopt a {lightweight memory} design, where the Coder is prompted round by round without retaining the full conversation history. 
In each round, it only receives the {\it latest} feedback from the Judge, previous kernel candidate, and task description, enabling it to focus on refining or correcting the previous version. 
This design greatly improves code stability, reduces inference cost, and allows the model to perform precise, hardware-driven optimizations. See Appendix~\ref{app:prompt} for Coder's prompt.
\vspace{-8pt}
\paragraph{Design of Correctness Tests.} 
After the Coder generates a candidate kernel, our workflow proceeds to verify its {\it correctness} 
through a two-stage {testing process}, including {compilation} and {execution}. 
The compilation stage ensures that the generated kernel is syntactically valid and can be successfully compiled into executable CUDA code. 
The execution stage then evaluates the functional correctness of the compiled kernel by running it on predefined test cases. 
We compare the kernel’s outputs with those from the PyTorch reference implementation under the same inputs, and consider the kernel {functionally correct} only if the numerical difference between the two outputs is within a small tolerance (set to $1\text{e}{-4}$ in our experiments). 
A kernel is regarded as {correct} only when it passes both compilation and execution stages successfully, which is a commonly adopted criterion~\citep{kevin-multi-turn-rl,lange2025robustagenticcudakernel,ouyang2025kernelbenchllmswriteefficient}.
\vspace{-8pt}
\paragraph{Design of Judge.} 
The {Judge} serves as the evaluation and guidance module in our framework. 
Its primary responsibility is to provide actionable feedback to guide the Coder’s next round given task description, current candidate kernel, and hardware feedback or runtime information. 
The Judge operates in two distinct modes depending on kernel validity:
If the kernel fails compilation or produces incorrect outputs, the Judge performs the {\it correction mode}, identifying issues such as invalid indexing, race conditions, or missing headers, and returns \textbf{correction feedback} that instructs the Coder to fix them. 
If the kernel passes all correctness checks, the Judge enters {\it  optimization mode}, where it uses hardware feedback  to identify the dominant bottleneck—e.g., memory-bound, compute-bound, or occupancy-limited—and formulates \textbf{optimization feedback} accordingly. Please see Section \ref{sec:searching} for details of using hardware feedback. 

Similar to the design of the Coder, the Judge does not retain the full conversation. 
In each round, the Judge is prompted with the relevant mode and acts according to the corresponding role. Finally, it generates structured feedback in JSON format, which is then extracted and passed to the Coder for the next round. 
This design ensures modularity and clarity in the interaction between agents, allowing the Judge to efficiently translate hardware-level profiling signals into precise, interpretable feedback that drives iterative kernel optimization. See Appendix~\ref{app:prompt} for the Judge's prompt.

\vspace{-12pt}

\subsection{How to Integrate Hardware Feedback}\label{sec:searching}
\vspace{-8pt}
In this subsection,  we describe in detail a key design consideration, which enables \agentname~to utilize hardware feedback for kernel performance optimization.
The {hardware feedback} integrates static GPU specifications (e.g. architecture, memory bandwidth, per-thread register limits, per-SM shared-memory capacity) with performance metrics (e.g. memory throughput, occupancy, and warp efficiency) from {Nsight Compute (NCU)} collected during kernel execution. By cross-referencing GPU specifications and NCU metrics, the Judge infers the kernel’s primary performance-limiting cause and bottleneck. Figure~\ref {fig:detail} illustrates how Judge uses hardware feedback to optimize kernels.

Just as CUDA engineers focus on key indicators, we choose not to pass the entire set of NCU metrics to the Judge. Feeding all metrics can overwhelm the decision process with excessive, partially redundant signals and lead to unstable judgments (See Section~\ref{NCU} and Appendix~\ref {app:why_subset} for ablation study and case study). Instead, we design a novel protocol which profiles a subset of critical metrics provided by NCU and forward them to Judge so that we can improve the quality of the judge outputs. 
More specifically, the key subset of metrics are selected off-line (before the workflow starts to work), through the following steps:

 \noindent{\bf (Step 1)} {\sf Kernel sampling and Selection}: We first profile key metrics on some preselected representative tasks (e.g., Conv2D, MatMul) to prepare a reliable metric set. Specifically, for each task we run 100 self-refine (repeating the cycle generating $\to$ execute/profile $\to$ evaluate $\to$ repair/optimize) with a single SOTA model (e.g. OpenAI-o3), collect the generated and correct kernels, and select $10$ with the largest speed disparity (fastest vs. slowest). See Algorithm~\ref{alg::alg1}. 
 
\begin{algorithm}[H]
\caption{Step 1: Kernel Sampling and Selection}\label{alg::alg1}
\KwIn{Task set $Task = \{T_1, T_2, \dots, T_n\}$}
\KwOut{Selected subsets $K_i^*$ for each task $T_i$}

\For{$i \gets 1$ \textbf{to} $n$}{
    $K_i \gets \emptyset$\;
    \For{$j \gets 1$ \textbf{to} $100$}{
        $k_j \gets \text{generate\_kernel}(T_i)$\;
        $K_i \gets K_i \cup \{k_j\}$\;
    }
    Sort $K_i$ in nondecreasing order according to kernel runtime\;
    $m \gets |K_i|$ \tcp*{Here $m=100$}
    $K_i^* \gets \{K_i[1], K_i[2], K_i[3], K_i[4], K_i[5], K_i[m-4], K_i[m-3], K_i[m-2], K_i[m-1], K_i[m]\}$\;
}
\end{algorithm}

  \noindent{\bf (Step 2)} {\sf Top-20 metrics within each task}: We then refine the metrics within each task to identify the most relevant candidates. Specifically, for each task we consolidate the NCU metrics from the $10$ kernels selected from Step 1 into a single dataset. Since Nsight Compute reports a consistent full set of metrics across all kernels, the metric categories are aligned by default. We then remove aliases and strongly collinear indicators, and compute Pearson correlations between each metric and kernel runtime. We retain only the Top-20 metrics (by absolute correlation) as the candidate set for this task (see Appendix~\ref{app:top-20}  for examples).

\noindent{\bf (Step 3)} {\sf Metrics selection across tasks}: Finally, we consolidate metrics across tasks to build a stable, task-agnostic set. We compare the Top-20 lists across tasks and keep metrics that consistently appear, show the same correlation direction, and achieve high global scores. Specifically, for each metric, we compute a {\it global correlation score} defined as the average of its absolute Pearson correlations with runtime across all tasks. We then select metrics whose global scores exceed the 75th percentile ($P_{75}$) among all candidates, ensuring that only the most strongly correlated metrics are retained. This yields 24 metrics that are strongly correlated with kernel runtime across tasks. Later, the Judge will profile each generated kernel with NCU and uses only this 24 metrics as references (see Appendix~\ref{app:key24} for the complete list of the selected metrics).  See Algorithm~\ref{alg::alg2}.

\begin{algorithm}[H]
\caption{Step 2-3: Profiling and Metrics Selection}\label{alg::alg2}
\KwIn{$K^* = \{K_1^*, K_2^*, \dots, K_n^*\}$, where each $K_i^* = \{k_1^*, k_2^*, \dots, k_{10}^*\}$}
\KwOut{Final metrics set $Final\_Metrics$}

$M^* \gets \emptyset$\;

\For{$i \gets 1$ \textbf{to} $n$}{
    $M_i^* \gets \emptyset$\;
    \ForEach{$k \in K_i^*$}{
        $M \gets \text{NCU\_Profile}(k)$ \tcp*{Run NCU profiling, $M=\{m_1, m_2, \dots, m_j\}$}
        \ForEach{$m \in M$}{
           $r_{m,i}$ = Compute Pearson correlation coefficient $r\!\left(m,\ \text{runtime}(k)\right)$\;
        }
        $Top20(k) \gets \text{the 20 metrics in } M \text{ with highest } |r(\cdot,\text{runtime}(k))|$\;
        $M_i^* \gets M_i^* \cup Top20(k)$\;
    }
    $M^* \gets M^* \cup M_i^*$\;
}
\tcp{Compute global correlation scores across tasks}
\ForEach{metric $m \in M^*$}{
    Compute $S_m = \frac{1}{n}\sum_{i=1}^{n} |r_{m,i}|$  
    \tcp*{$r_{m,i}$: Pearson correlation between metric $m$ and runtime on task $i$}
}

\tcp{Select stable and highly correlated metrics}
$Final\_Metrics \gets \{m \mid m \text{ appears in multiple tasks, keeps same sign, and } S_m > P_{75}(S)\}$\;

\tcp{Final set contains 24 distinct metrics}
\end{algorithm}

After the key subset of NCU metrics is determined offline, the Judge will use these metrics to identify performance bottlenecks in the \agentname~workflow. At each optimization round, the Judge profiles the generated kernel with NCU and collects hardware feedback, including static GPU specifications and the key subset of NCU metrics.  Based on this information, the Judge identifies the dominant bottleneck in the current kernel. 
To prevent AI agent reasoning without direction and generating suboptimal results, the Judge is prompted to only capture 3-4 most important metrics in each round according to its own reasoning. For example, Judge can identify the current kernel as memory-bound when memory throughput is high but computing resource utilization is low, and then it will choose memory related metrics as critical metrics in this round. After this, Judge will generate suggestions on how to modify the kernel to address the current critical bottleneck. The Coder incorporates this guidance in the next round generation accordingly. This mechanism enables our multi-agent system focus on addressing only one critical program bottleneck in each round, and eventually optimizes overall kernel performance step by step in iterative rounds, just like human experts' real workflow.

\section{Experiments}\label{exp}
\subsection{Benchmark and Evaluation}
We evaluate our method and baselines on \textbf{KernelBench}~\citep{ouyang2025kernelbenchllmswriteefficient}, a popular benchmark designed to assess the ability of LLMs to generate CUDA kernels.  
KernelBench consists of multiple difficulty levels, and we adopt all tasks from Level 1 to Level 3, resulting in a total of 250 tasks.  
Specifically, Level 1 contains relatively simple 100 tasks involving basic operators (e.g., matrix multiplication),  
Level 2 includes medium-difficulty 100 tasks composed of multi-step operator combinations,  
and Level 3 contains 50 challenging tasks involving full neural network architectures (e.g., AlexNet).  
Each task is accompanied by a reference PyTorch implementation and predefined input/output specifications, which enables fully automated and reliable evaluation of both correctness and performance. Details of KernelBench are provided in Appendix~\ref{KernelBench}. 

Due to the high computational cost of running experiments for some ablation studies, we do not evaluate them on the entire KernelBench benchmark. 
Instead, we construct a {stratified random subset} of tasks, denoted as $\mathcal{D}^{*}$, by sampling proportionally from each difficulty level of KernelBench. This ensures that $\mathcal{D}^{*}$ maintains the same task distribution as the full benchmark while enabling fair and efficient evaluation. Specifically, $\mathcal{D}^{*}$ contains a total of 25 tasks, with 10 tasks in Level 1, 10 tasks in Level 2 and 5 tasks in Level 3. More details on the construction of $\mathcal{D}^{*}$ are provided in Appendix~\ref{subset}.

We evaluate model performance on KernelBench using the following metrics: (1) \textbf{Correctness}: the fraction of tasks for which the generated kernel compiles successfully and produces outputs identical to the PyTorch reference on all test cases. (2) \textbf{Performance}: the ratio of the execution speed (tested on a specific GPU), between a correct generated kernel and its PyTorch reference. 
(3) \textbf{Fast\(_1\)} : the proportion of correct kernels whose execution speed exceeds their PyTorch reference. 
(4) \textbf{Median speedup}: the median of `Performance' values across all tasks, reflecting typical rather than average behavior.
(5) \textbf{75th percentile speedup}: the 75th percentile of Performance values, capturing upper-quartile efficiency.

For methods that perform iterative refinement or generate multiple candidates (including \agentname), we report the best-performing correct kernel among all candidates for each task.

\subsection{Settings \& Baselines}
In our main results, we instantiate \agentname~with OpenAI-o3 as both the Coder and the Judge as our {\it default} setting.  
We set the maximum number of iteration rounds to $N{=}10$ to balance performance improvements and inference cost.  
Unless otherwise stated, all methods are evaluated under the same compilation/runtime environment in Quadro RTX 6000 and task-specific test suites. 

To contextualize the performance of \agentname~and assess the effect of advanced foundation models, we include the following baselines for the main results and ablation studies:

\begin{itemize}

    \item OpenAI-o3: Using OpenAI-o3 for one-shot generation without iteration;  
    \item o3-self-refine(our baseline): Using OpenAI-o3 for ten rounds of self-refinement without a Judge, where the model relies solely on itself to correct and optimize kernels given hardware feedback;  
    \item o3-correction(our baseline): A variant of \agentname~where the Judge provides only correctness feedback but no optimization feedback;  
    \item o3-optimization(our baseline): A variant of \agentname~where the Judge provides only optimization feedback but no correction feedback;  
    \item Kevin-32B: a strong RL-based model for CUDA kernel optimization from \citep{kevin-multi-turn-rl}. We directly take results from their official paper.  
    \item Agentic Baseline: the agentic workflow from \citep{lange2025robustagenticcudakernel}, a strong multi-agent baseline, which uses an LLM ensemble including both
reasoning (o3 \& o4-mini) and conventional LLMs (Claude Sonnet 3.7 \& GPT-4.1).
    \item \agentname~(full metrics): a variant of \agentname~where the Judge leverages the entire set of NCU metrics.
\end{itemize}

This baseline setting enables a comprehensive comparison across (i) base model vs.\ corresponding agent-based method, (ii) the presence/absence of Judge feedback, (iii) RL-based vs.\ training-free agent-based approaches, and (iv) different agentic methods. 

\begin{table}[t]
\caption{Main results on KernelBench (Level 1-3, 250 tasks). Results of Agentic Baseline is on Level 1 and 2. All experiments here are run in RTX 6000. Methods evaluated on \(\mathcal{D}^{*}\) are marked with \(^*\). \agentname-Scaling Up means scaling up maximum iteration rounds}

\label{tab:main_results}
\centering
\begin{tabular}{lcccccc}
\toprule
\textbf{Method} & \textbf{Correct$\uparrow$} & \textbf{Median $\uparrow$}  & \textbf{75\% $\uparrow$}  & \textbf{Perf $\uparrow$}  & \textbf{Fast$_1$$\uparrow$} \\
\midrule
OpenAI-o3         & 57.6\% & 0.390 & 1.014 & 0.680  & 31.60\% \\
o3-self-refine        & 90.8\% & 1.012 & 1.209 & 1.107  & 55.20\% \\
o3-correction      & 97.6\% & 1.031 & 1.238 & 1.222  & 59.60\% \\
o3-optimization      & 88.4\% & 1.061 & 1.483 & 1.509  & 64.00\% \\
Agentic Baseline(Level 1 \& 2) & 95.0\% & --- & ---  & 1.490 & --- \\
\agentname(full metrics)$^{*}$   & 100\% & 1.280 & 1.489  & 1.414 & 80.00\% \\
 
\midrule
\agentname  & \textbf{97.6\%} & \textbf{1.107} & \textbf{1.592} & \textbf{1.677}  & \textbf{70.80\%} \\

\agentname(Level 1 \& 2)  & \textbf{98.0\%} & \textbf{1.112} & \textbf{1.617} & \textbf{1.776}  & \textbf{71.50\%} \\

\agentname$^{*}$ & \textbf{100\%} &  \textbf{1.322} & \textbf{1.736} & \textbf{1.767} & \textbf{84.00\%} \\
\agentname-Scaling Up$^{*}$ & \textbf{100\%} &  \textbf{1.317} & \textbf{1.777} & \textbf{2.265} & \textbf{92.00\%} \\
\bottomrule
\end{tabular}
\end{table}

\begin{table}[t]
\caption{Main results on KernelBench (Level 1-3, 250 tasks) of \agentname~ in RTX 6000.}
\label{tab:Level1-3}
\centering
\begin{tabular}{lccccc}
\toprule
\textbf{Task} & \textbf{Correct$\uparrow$} & \textbf{Median $\uparrow$}  & \textbf{75\% $\uparrow$}  & \textbf{Perf $\uparrow$}  & \textbf{Fast$_1$$\uparrow$} \\
\midrule
Level 1   & 96\% &  1.044 & 1.751 & 1.448 & 54.0\% \\
Level 2  & 100\% & 1.124 & 1.427 & 2.104 & 89.0\% \\
Level 3  & 96\% & 1.081& 1.510 & 1.283 & 68.0\% \\
\bottomrule
\end{tabular}
\end{table}

\subsection{Main Results in RTX 6000}
\label{sec:MainResults}
Table~\ref{tab:main_results} reports the main results in KernelBench. {\agentname} consistently outperforms all baselines across all metrics, both in the entire Kernelbench and in the stratified subset \(\mathcal{D}^{*}\) .

{On KernelBench, \agentname~attains \textbf{97.6\%} correctness with an average performance of \textbf{1.677$\times$}, and \textbf{70.8\% Fast$_1$}, while achieving a median speedup of 1.107$\times$ with a 75th percentile speedup of 1.592$\times$. These results significantly improve over the base model OpenAI-o3 and other ablated variants, including o3-self-refine, o3-correction and o3-optimization.}

{On the reduced dataset \(\mathcal{D}^{*}\), \agentname~achieves 100\% correctness, a median speedup of 1.322$\times$, a 75th percentile speedup of 1.736$\times$, an average performance of 1.767$\times$, and 84.0\% Fast$_1$.  
This substantially surpasses \agentname(full metrics), which reaches only 1.280$\times$ median, 1.489$\times$ at the 75th percentile, 1.414$\times$ performance, and 80\% Fast$_1$. Moreover, after scaling up maximum iteration rounds, \agentname-Scaling Up achieves a stronger performance, with \textbf{2.265$\times$} speedup. We will further discuss ablation studies using  \(\mathcal{D}^{*}\) in Section \ref{ablation} and scaling up maximum iteration rounds in Section \ref{scalingN}.}

\begin{figure}[t]
    \centering
    \includegraphics[width=0.85\linewidth]{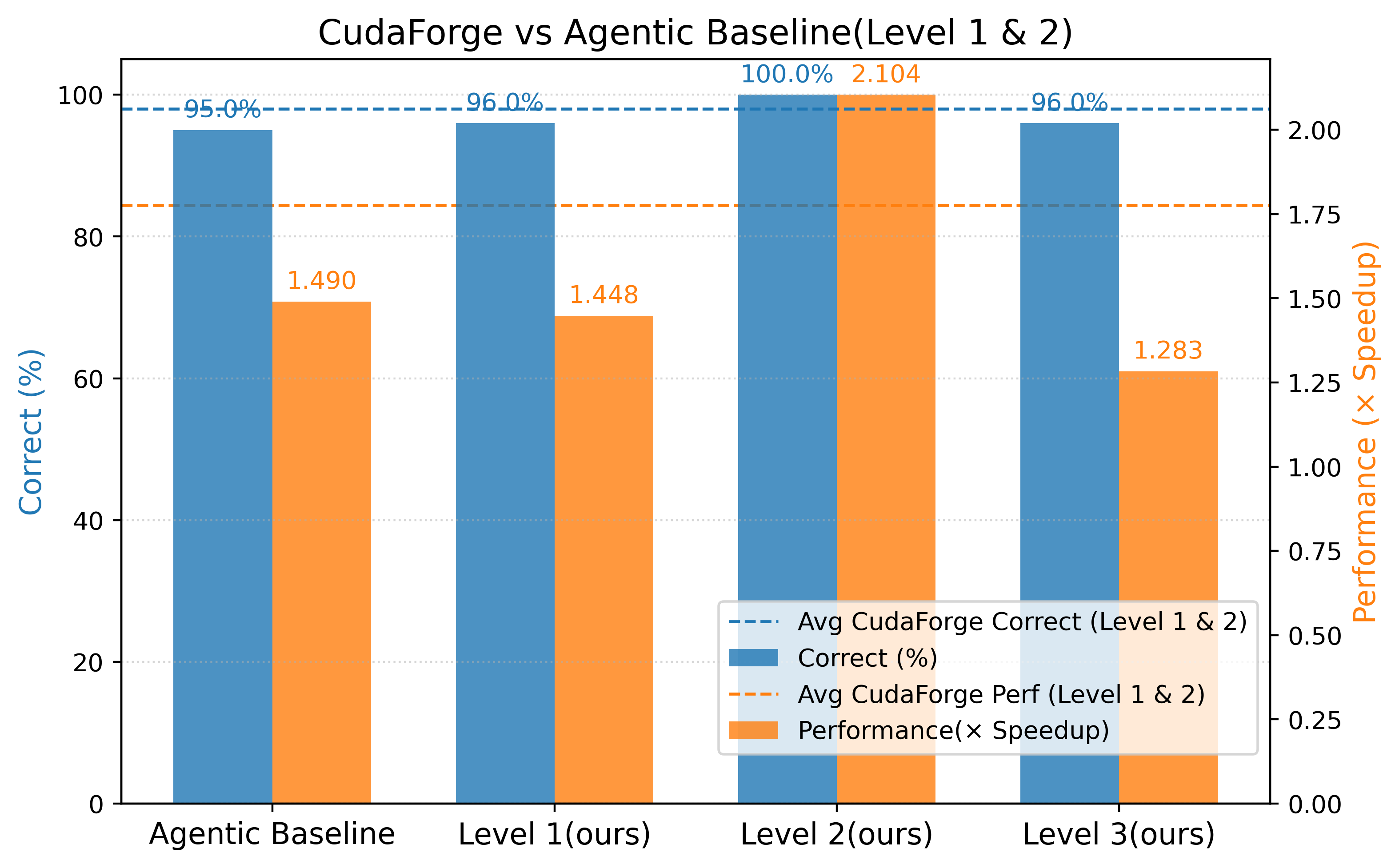}
    \caption{
        Comparison of correctness and performance between \agentname~and the Agentic Baseline on KernelBench. Dashed lines denote average results of \agentname~over Level 1 and 2. \agentname~outperforms Agentic Baseline on KernelBench Level 1 and 2, and it also achieves strong performance in Level 3.
    }
    \label{fig:cuda_benchmark}
\end{figure}

We also compare \agentname~with Agentic Baseline\footnote{Note that it only reports results in Level 1 and 2, and we directly take the results from their paper since the paper has not open-sourced the code.} on KernelBench Level 1 and Level 2. As shown in Table~\ref{tab:main_results},~\ref{tab:Level1-3} and Fig~\ref{fig:cuda_benchmark}, \agentname ~achieves 98\% correctness and an average speedup of 1.776$\times$, which outperforms Agentic Baseline (95.0\%, 1.490$\times$), especially in speedup. This result shows our advantage compared to existing agentic work.

Notably, on Level 3, which represents the most challenging level of KernelBench, \agentname~achieves \textbf{96\%} correctness and an average \textbf{1.283$\times$} speedup.  
Given the complexity of Level 3 tasks, which involve full neural network architectures and multi-stage operations, these results demonstrate that \agentname~is capable of reliably generating and optimizing highly complex CUDA kernels, where prior approaches~\citep{kevin-multi-turn-rl,lange2025robustagenticcudakernel} have not explored it.

\subsection{{Comparison with Kevin-32B on H200}}
\label{Kevin}

In this subsection, we compare \agentname~with Kevin-32B~\citep{kevin-multi-turn-rl}. However, since Kevin-32B is not open-sourced—meaning that key details such as test-time prompts, evaluation setup, and benchmark specifications are not publicly available—a fully fair and reproducible comparison cannot be conducted.
According to the original paper, Kevin-32B was trained and evaluated on H200 GPUs using a self-constructed benchmark of comparable difficulty to KernelBench Levels 1 and 2. To ensure the comparison is as fair as possible under these constraints, we re-evaluate \agentname~on the same H200 hardware across KernelBench Levels 1–3,  aligning as much as possible our computational environment and evaluation protocol with those reported for Kevin-32B. 

\begin{figure}[t]
    \centering
    \includegraphics[width=0.85\linewidth]{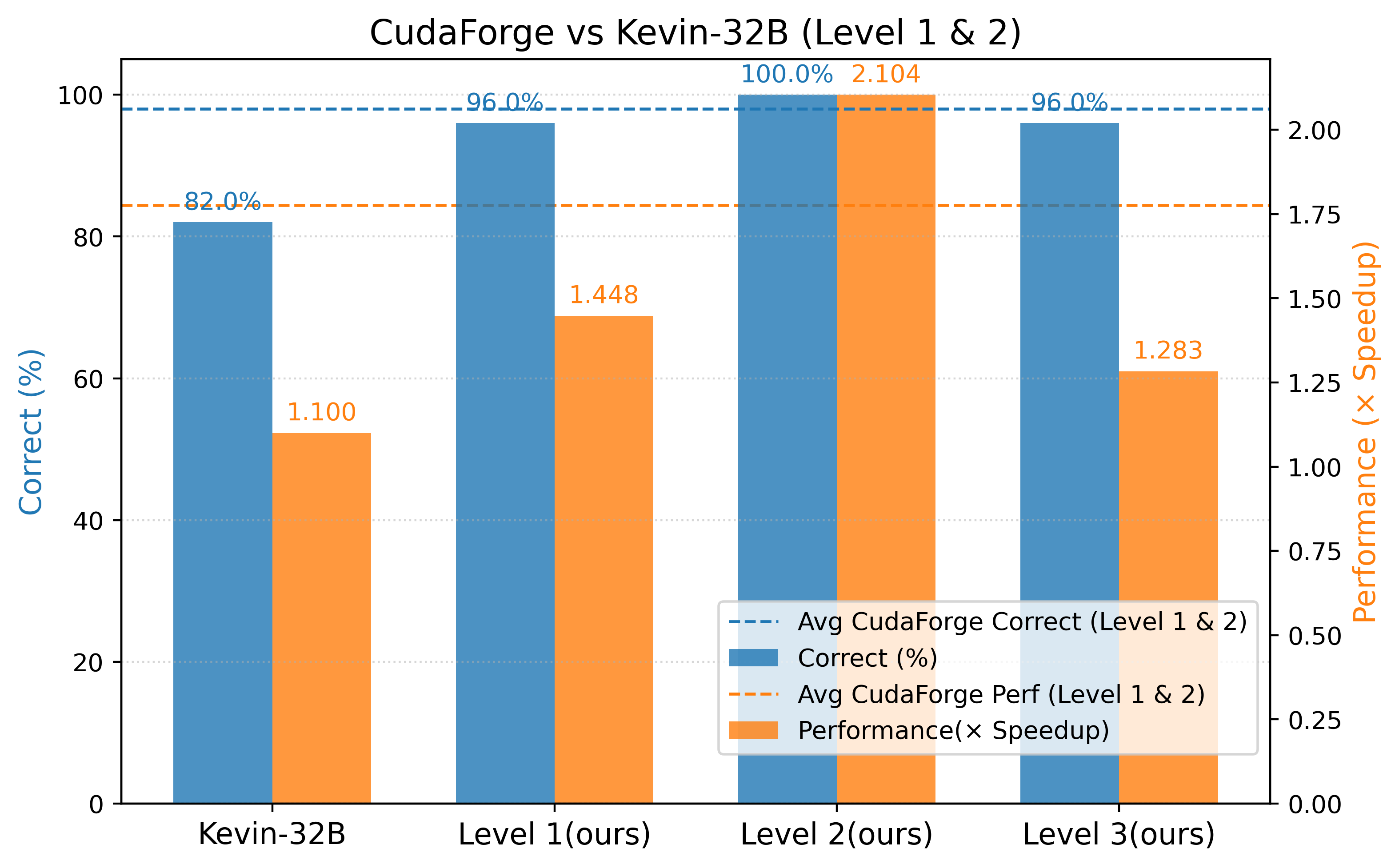}
    \caption{
        Comparison of correctness and performance between \agentname~and Kevin-32B on KernelBench. Dashed lines denote average results of \agentname~over Level 1 and 2. While training-free, \agentname~outperforms Kevin-32B in KernelBench Level 1-2, and gets outstanding results in Level 3.
    }
    \label{fig:kevin}
\end{figure}

As shown in Figure~\ref{fig:kevin}, \agentname~achieves consistently higher correctness and performance across all levels.
In the tasks comparable to Kevin’s benchmark (Levels 1 \& 2), \agentname~reaches an average of \textbf{98.0\%} correctness and \textbf{1.662$\times$} performance, outperforming Kevin-32B’s {82.0\% correctness} and {1.10$\times$ speedup}.
Even in the most challenging Level 3 tasks, \agentname~maintains strong results with \textbf{96.0\%} correctness and \textbf{1.261$\times$} performance, demonstrating its robustness on complex CUDA kernels.

These findings highlight that, despite being a training-free framework, \agentname~surpasses the RL-trained Kevin-32B in both reliability and efficiency.
We attribute this advantage to the effective use of hardware feedback and the design of workflow, which enables \agentname~to perform targeted hardware-aware optimizations rather than rely solely on (potentially inefficient) RL training.

\subsection{API and Computation Time Cost of \agentname}
We evaluate both the API and computation time costs of \agentname~on the KernelBench dataset~$\mathcal{D^*}$. 
The {API cost} is measured as the total expenditure per kernel generation task, 
while the {computation time} is measured as the end-to-end wall-clock time, 
including kernel compilation and execution, model generation, and NCU profiling. 
All experiments are conducted on a single RTX~6000 GPU.  

\begin{table}[t]
\caption{
Comparison of API and computation time cost between \agentname~and the Agentic Baseline~\citep{lange2025robustagenticcudakernel}.
}
\label{tab:cost}
\centering
\begin{tabular}{lccccc}
\toprule
\textbf{Method} & \textbf{Metric} & \textbf{Average} & \textbf{Level 1} & \textbf{Level 2} & \textbf{Level 3} \\
\midrule
\multirow{2}{*}{Agentic Baseline} 
 & API Cost (\$) & 5.0 & --- & --- & --- \\
 & Time (min) & 60.0 & --- & --- & --- \\
\midrule
\multirow{2}{*}{\agentname} 
 & API Cost (\$) & \textbf{0.30} & \textbf{0.29} & \textbf{0.30} & \textbf{0.33} \\
 & Time (min) & \textbf{26.5} & \textbf{28.5} & \textbf{24.1} & \textbf{27.1} \\
\bottomrule
\end{tabular}
\vspace{-0.5em}
\end{table}

As shown in Table~\ref{tab:cost}, \agentname~requires on average only {26.5 minutes} of wall-clock time on a single RTX~6000 GPU and incurs merely {\$0.3} of API cost per kernel. 
This is highly cost-efficient compared with the Agentic Baseline~\citep{lange2025robustagenticcudakernel}, 
which reports approximately {6 GPU hours on H100} and {\$5} per kernel in their Appendix~E.  

\begin{figure}[t]
\centering
\begin{subfigure}[t]{0.48\textwidth}
    \centering
    \includegraphics[width=\linewidth]{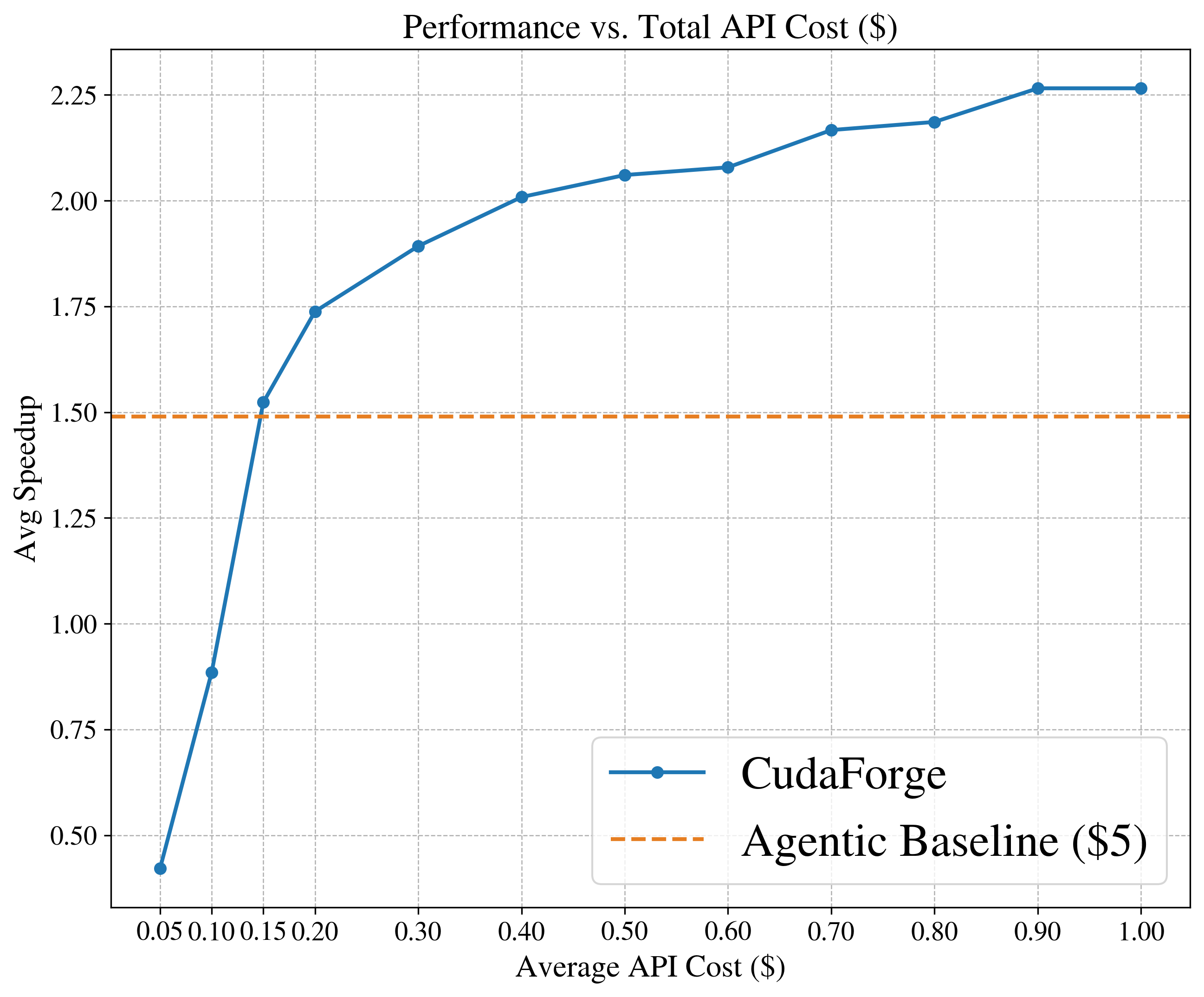}
    \caption{Performance vs. API cost of \agentname.}
    \label{fig:api_cost}
\end{subfigure}
\hfill
\begin{subfigure}[t]{0.48\textwidth}
    \centering
    \includegraphics[width=\linewidth]{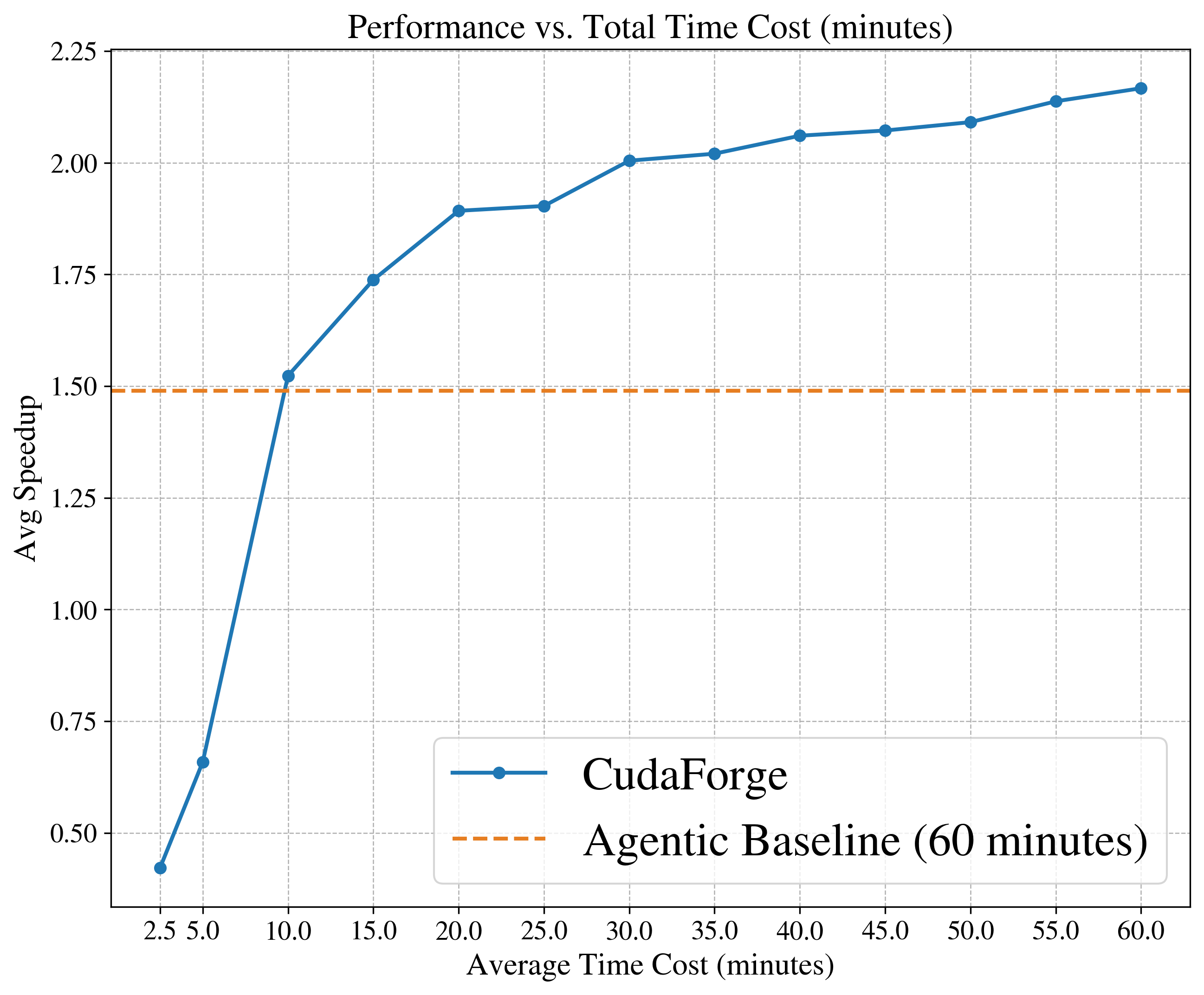}
    \caption{Performance vs. computation time of \agentname.}
    \label{fig:time_cost}
\end{subfigure}
\vspace{-0.5em}
\caption{
Relationship between cost and performance of \agentname.
Both API and computation time exhibit a monotonic correlation with performance.
\agentname~already surpasses the Agentic baseline while using no more than {\$0.15} and {10 minutes} per task.
}
\label{fig:apicost_timecost}
\vspace{-0.5em}
\end{figure}

We further analyze how the API and computation costs relate to \agentname's performance, as illustrated in Figure~\ref{fig:api_cost} and~\ref{fig:time_cost}. 
The performance of \agentname~monotonically improves as the API and computation time increase. 
Notably, \agentname~already surpasses the Agentic baseline~\citep{lange2025robustagenticcudakernel} while using no more than {\$0.15} and {10 minutes} per task, 
demonstrating its strong cost–performance tradeoff.

We attribute \agentname's cost efficiency to three main factors:
\begin{enumerate}
    \item \textbf{leveraging hardware feedback.}  
    The Judge leverages hardware feedback to diagnose the current bottleneck and provide targeted, actionable optimization guidance to the Coder. 
    This focused refinement avoids blind exploration and accelerates convergence, thereby reducing the number of API calls and profiling rounds.  

    \item \textbf{Selective NCU metrics.}  
    Instead of profiling the entire metric set, \agentname~uses a curated subset of critical NCU metrics. 
    This not only shortens NCU profiling time but also reduces API cost, as fewer metrics decrease the input token length in Judge queries. 
    \item \textbf{Lightweight memory.}
    In each iteration, both the Coder and Judge are prompted with current round's information, instead of using the full conversation history. 
    This lightweight memory design minimizes redundant context tokens and computation overhead, allowing each agent to focus solely on the most recent feedback and candidate kernel. 
\end{enumerate}

\subsection{Ablation Studies}
\label{ablation}
\paragraph{{Comparison with using the entire set of NCU metrics.}}
\label{NCU}
A key design choice in \agentname~is to filter the full set of NCU metrics and retain only a subset of 24 critical metrics for the Judge.
This selective design allows the Judge to focus on the most informative performance indicators, avoiding redundancy and enabling more consistent optimization feedback.
We conduct an ablation study to evaluate this choice.
As shown in Table~\ref{tab:main_results}, using the complete set of NCU metrics leads to lower correctness and performance, as the Judge is overwhelmed by excessive, partially redundant signals.
Moreover, profiling with all NCU metrics substantially increases inference cost—each kernel requires approximately 40 miuntes on an RTX 6000 GPU and incurs about \$1 in API cost—whereas our selective-metric design achieves superior performance with far lower overhead(26.5 minutes on an RTX 6000 and \$0.3 in API cost).
These results confirm that concise, focused hardware feedback is both more effective and more efficient than exhaustive profiling. We also provide a case study in Appendix~\ref {app:why_subset}.

\paragraph{Comparison with o3-self-refine.}  
A key motivation behind \agentname~is to decouple the roles of generation and evaluation.  
In o3-self-refine, the same model performs ten rounds of self-refinement, implicitly taking on both roles: it must both propose new kernels and evaluate its own outputs based on hardware feedback and runtime signals.  
While this strategy improves the correctness percentage from 57.6\%  to 92.8\%, performance remains limited ({1.107$\times$} speedup, {55.2\% Fast$_1$}).  
In contrast, \agentname~explicitly separates responsibilities: the Coder focuses on code generation, while the Judge specializes in providing structured feedback.  
This division of labor proves critical—allowing each agent to concentrate on a distinct reasoning process—and results in significantly higher efficiency ({1.677$\times$} speedup, {70.8\% Fast$_1$}) without sacrificing correctness.
\label{Ablation-agents}
\paragraph{Comparison with o3-correction (correction-only Judge).}  
In o3-correction, the Judge only provides correction feedback based on runtime signals, without optimization feedback.  
This setting achieves the same {97.6\% correctness} as \agentname, confirming that iterative error correction is sufficient to ensure reliable kernel generation.  
However, efficiency remains much lower, with only {1.222$\times$} performance and {58.8\% Fast$_1$}.  
The contrast with \agentname (\textbf{1.677$\times$}, \textbf{70.8\%}) highlights that while correctness feedback stabilizes generation, performance feedback (grounded in hardware profiling) is essential for driving substantial efficiency gains.

\paragraph{Comparison with o3-optimization (optimization-only Judge).}  
We also evaluate the variant where the Judge provides only optimization feedback, without correction feedback.  
In this setting, the Coder frequently generates kernels that fail to compile or run, since functional errors remain uncorrected.  
As a result, overall correctness is substantially lower than \agentname, and the potential benefits of optimization guidance cannot be realized.  
This outcome demonstrates that correctness feedback is a prerequisite: Without first ensuring functional validity, optimization feedback alone is ineffective and often wasted.  
In contrast, \agentname~leverages both correction and optimization feedback, enabling stable kernel generation and consistent efficiency improvements.

\subsection{Generalization Capability of \agentname}
In this section, we analyze \agentname's capabilities across various maximum iteration num $N$, GPU architectures and base models. Considering the high cost of full experiment, we use the stratified subset \(\mathcal{D}^{*}\) for this section.

\noindent{\bf Scaling up the maximum number of iteration rounds.}\label{scalingN}
We investigate the effect of the maximum iteration number $N$ on \agentname’s performance. 

\begin{figure}[hbp]
  \centering
  \includegraphics[width=0.6\linewidth]{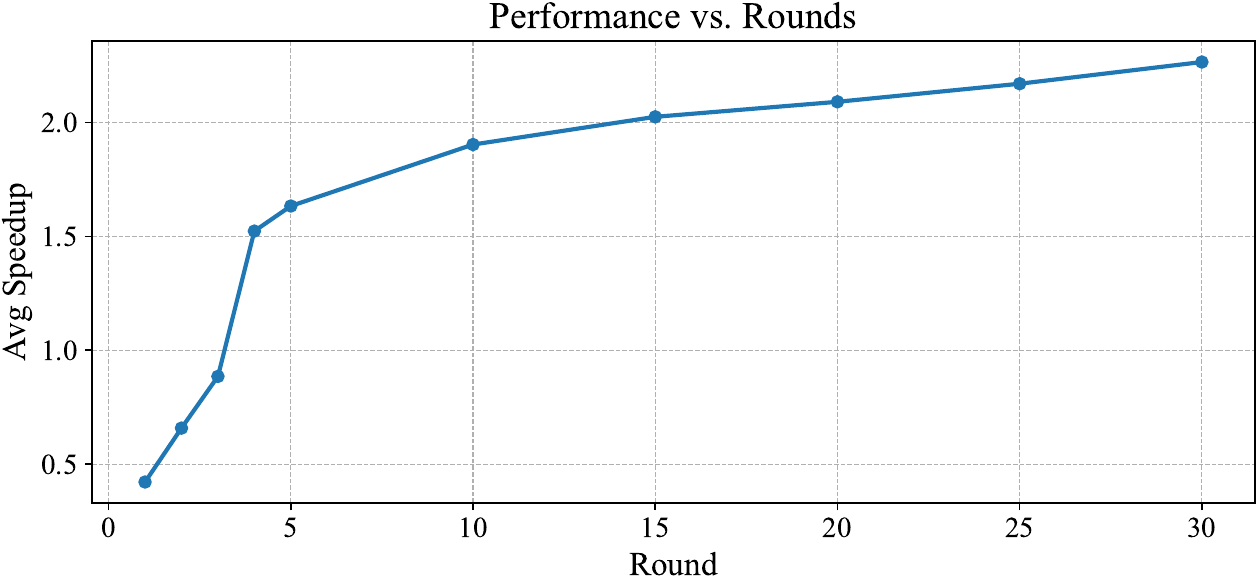}
  \caption{Scaling the number of iteration rounds to 30 on KernelBench (subset $\mathcal{D}^{*}$).}
  \label{fig:scaling_N}
\end{figure}

As shown in Figure~\ref{fig:scaling_N}, increasing $N$ from 1 to 10 leads to substantial performance gains, indicating that \agentname~can rapidly improve kernel efficiency through iterative refinement. 
Further increasing $N$ from 10 to 30 continues to improve performance, though with a slower growth rate, suggesting that the system gradually approaches its performance ceiling. After 30 rounds of optimization, \agentname~increases the average speedup to 2.271$\times$.
These results demonstrate that \agentname~benefits from scaling up and has the potential to achieve even stronger performance given larger $N$ with additional inference cost.

\noindent{\bf Using \agentname~in different GPUs.}
\begin{table}[t]
\caption{\agentname's performance on different GPUs. The system consistently achieves high correctness and strong performance across architectures by incorporating GPU specifications and \textit{Nsight Compute} profiling signals during optimization.}
\label{tab:cross_gpu}
\centering
\begin{tabular}{lccccc}
\toprule
\textbf{GPU} & \textbf{Correct$\uparrow$} & \textbf{Median $\uparrow$}  & \textbf{75\% $\uparrow$}  & \textbf{Perf $\uparrow$}  & \textbf{Fast$_1$$\uparrow$} \\
\midrule
RTX 6000(Ada Arch-Data center level)   & 100\% &  1.322 & 1.736 & 1.767 & 84.0\% \\
RTX 4090(Ada Arch-Desktop level)  & 100\% & 1.188 & 1.589 & 1.327 & 80.0\% \\
A100(Ampere Arch-Data center level)  & 100\% & 1.371& 1.762 & 1.841 & 84.0\% \\
RTX 3090(Ampere Arch-Desktop level)  & 100\% & 1.155& 1.706 & 1.320 & 72.0\% \\
\bottomrule
\end{tabular}
\end{table}
We also evaluate \agentname~ on various GPU architectures, including RTX 6000, RTX 4090, RTX 3090 and A100, to examine its effectiveness under different hardware conditions.  
As shown in Table~\ref{tab:cross_gpu}, \agentname
~consistently achieves high correctness and strong performance on all tested GPUs. This is a direct consequence of its design: during the optimization phase, the Judge explicitly incorporates hardware feedback, including NCU metrics and GPU specifications when generating feedback to Coder. This allows the Coder to produce kernels that are tailored to the target GPU at inference time, without training.

\noindent{\bf Instantiate \agentname~with various LLM.}
\begin{table}[t]
\caption{Performance of \agentname~with different base model combinations. We fix one agent as OpenAI-o3(denoted as O3) and replace the other with various models. All combinations achieve strong results, showing that the framework is not tied to a specific base model.} 
\label{tab:base_models}
\centering
\begin{tabular}{lccccc}
\toprule
\textbf{Models (Coder/Judge)} & \textbf{Correct$\uparrow$} & \textbf{Median $\uparrow$}  & \textbf{75\% $\uparrow$}  & \textbf{Perf $\uparrow$}  & \textbf{Fast$_1$$\uparrow$} \\
\midrule
O3 / O3            & 100\% &  1.322 & 1.736 & 1.767 & 84.0\% \\
\midrule
O3 / GPT-5          & 100\% & 1.131 & 1.561 & 2.114 & 96.0\% \\
O3 / Claude-Sonnet-4         & 100\% & 1.265 & 1.456 & 1.829 & 84.0\% \\
O3 / GPT-OSS-120B    & 100\% & 1.226 & 1.490 & 1.364 & 76.0\% \\
\midrule
GPT-5 / O3           & 100\% & 1.125 & 1.388 & 1.896 & 72.0\% \\
Claude-Sonnet-4 / O3          & 88\% & 1.052 & 1.207 & 1.398 & 56.0\% \\
GPT-OSS-120B / O3    & 96\% & 1.080 & 1.477 & 1.653 & 68.0\% \\
QwQ / O3   & 84\% & 0.965 & 1.153  & 0.790 & 44.0\% \\
\bottomrule
\end{tabular}
\end{table}
To examine whether \agentname~depends on a specific base model, we conduct experiments by fixing one side (Coder or Judge) as \textit{OpenAI-o3}(denoted as O3) and replacing the other with various advanced LLMs, including \textit{QwQ-32B}, \textit{GPT-5}, \textit{Claude-Sonnet-4}, and \textit{GPT-OSS-120B}.  
As shown in Table~\ref{tab:base_models}, all combinations achieve high correctness and strong performance, comparable to or even surpassing the original O3/O3 configuration.  
These results indicate that \agentname~is not tied to a specific base model: its effectiveness stems from the workflow of Coder and Judge, and it can readily benefit from stronger models as they emerge.

\section{Case study}
\begin{figure}[htbp]
  \centering
  \includegraphics[width=1.0\linewidth]{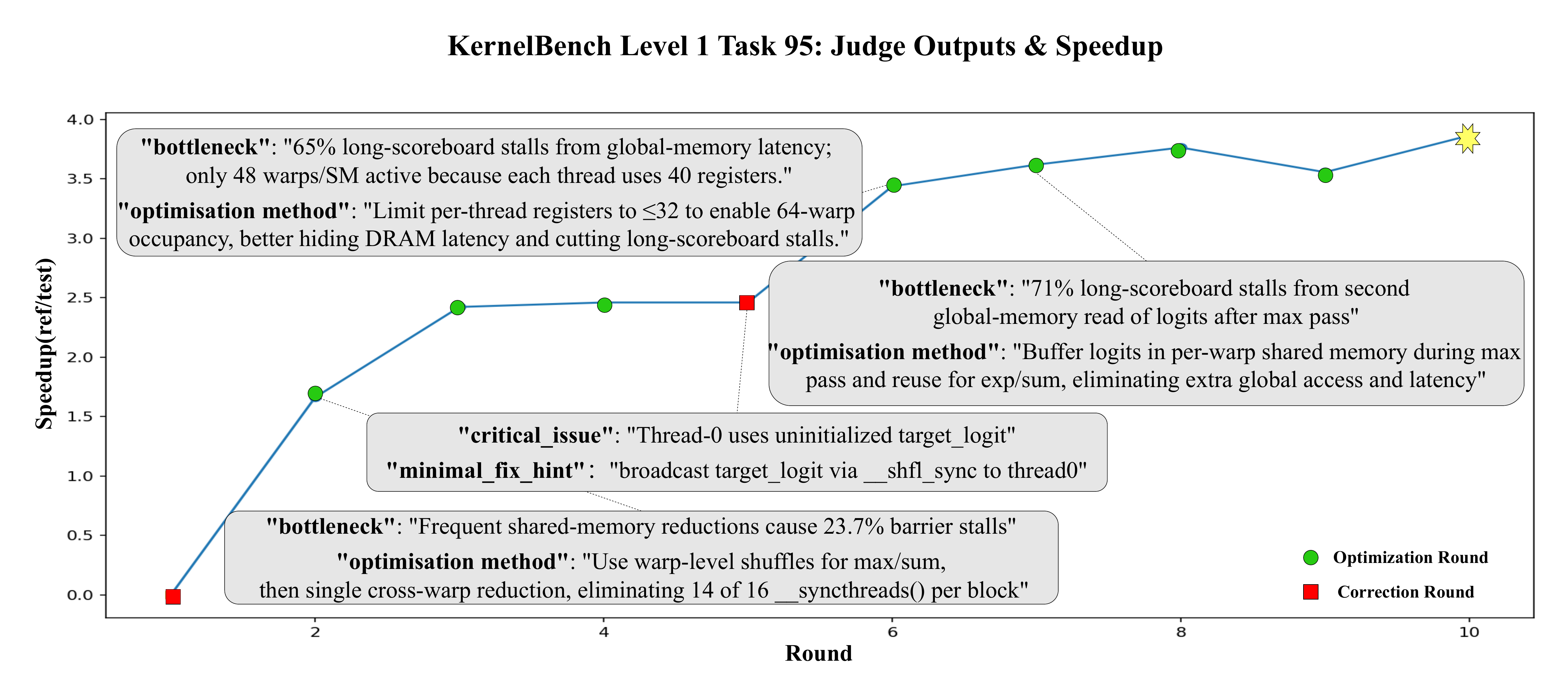}
  \caption{Illustration of the Judge’s outputs—bottleneck diagnoses and optimization suggestions—on KernelBench Level-1 Task 95 (CrossEntropyLoss), as well as the correspondi speedup across rounds (green = optimization, red = correction).}
  \label{fig:good_case}
\end{figure}

In this section, we present a case study on a single task to illustrate how the Judge diagnoses issues and recommends optimizations. Figure~\ref{fig:good_case} depicts the 10-round refinement process of \agentname~ on task \texttt{95\_CrossEntropyLoss}. We highlight four representative rounds—three optimization rounds and one repair round—to demonstrate how the Judge leverages hardware feedback from NCU to provide targeted optimization or bug-fix suggestions.

In round 2, which is an optimization round, the Judge notices that 23.7\% of active warps are stalled due to barrier-type dependencies, which means roughly one quarter of potential issue opportunities are blocked by synchronization. According to this, the Judge recommended replacing the original shared-memory reduction that required multiple block-level synchronizations with a warp-level shuffle reduction, giving below suggestion as prompt for coder: use warp-level shuffles in the max and sum phases, then perform a single cross-warp combine, reducing \texttt{\_\_syncthreads()} per block from 16 to 2 (a reduction of 14). After applying this change, performance improved from \(\mathbf{1.66\times}\) to \(\mathbf{2.42\times}\), with barrier stalls reduced and instruction-issue efficiency increased.

In round 5, it is a correction round. The previous round fails a numerical check with the following error: ``Outputs are not close, indicating a result mismatch". The Judge diagnosed the root cause as an uninitialized target\_logit in thread 0 (“Thread-0 uses uninitialized target\_logit”), which means the variable target\_logit is not updated to thread 0, leading wrong computing results. Accordingly, the Judge gave the minimal fix suggestion, broadcast \texttt{target\_logit} via \texttt{\_shfl\_sync} to thread 0. After applying the fix, the numerical issue disappeared.

In rounds 6 \& 7 (both optimization rounds), the Judge continues to track smsp\_warp\_issue \_stalled\_long\_scoreboard\_per\_warp\_active.pct. 
In round~6, this metric is about \(65\%\), primarily reflecting long-scoreboard stalls caused by global-memory latency. Per-thread register usage is high, resulting in limited occupancy (only \(\sim48\) active warps/SM) and insufficient latency hiding. 
The recommendation is to reduce per-thread registers to raise concurrency to \(\sim64\) warps/SM and thereby lower the long-scoreboard share. 
In round~7, the metric rises to about \(71\%\), rooted in a second global read of logits after the max pass. 
The Judge therefore advises buffering logits in per-warp shared memory during the max pass and reusing them in the exp\/sum phase, eliminating the redundant global memory access. 
Together, these strategies reduce global memory access, significantly cut long-scoreboard stalls, improve issue efficiency and throughput; after these two rounds, the speedup increases from \(3.436\times\) to \(3.762\times\).

This task demonstrates our \agentname's stability and expert-like workflow: first analyzing bottlenecks from hardware feedback, then deriving the corresponding optimization strategy.

\section{Supplement Observations}

\noindent{\bf Observations in CUDA-L1 results.} We carefully examined the kernel outputs reported by CUDA-L1 (see Appendix~\ref{app:cuda-l1}) and identified an interesting phenomenon that we term \emph{"fake kernels".}
These kernels, while reported as performant, often contain no actual CUDA code. Instead, they rely on \texttt{try-except} constructs and fall back to PyTorch’s official implementations to solve the task.
This observation highlights a fundamental challenge in evaluating LLM-generated CUDA kernels. To avoid this issue, we have manually checked all kernels in our experiments.

\section{Conclusion}
We presented \agentname, a training-free multi-agent framework for CUDA kernel generation and optimization.
The framework imitates the iterative workflow of human experts, explicitly incorporating hardware feedback to guide targeted kernel refinement rather than blind exploration. On the KernelBench benchmark, \agentname~achieves highest correctness rate and significant performance gains compared with all existing method, while also demonstrating robustness across diverse GPU architectures and base LLMs 
Moreover, its performance scales effectively with the number of refinement rounds.
Finally, thanks to its low API and time cost, \agentname~provides a practical and efficient solution for automated CUDA kernel development.

\newpage
{{
\bibliographystyle{unsrtnat}
\bibliography{reasoning}
}}

\clearpage
\newpage
\appendix
\section{Description of the Prompts used in \agentname~Workflow}\label{app:prompt}
\subsection{Prompt for Coder in first round of generation} 
We adopt the \emph{One-shot Baseline Prompt} introduced in \textsc{KernelBench} as our initial prompt for first round generation of all the baselines and our method. The full prompt is shown below.

\begin{tcolorbox}[mybox, title={Prompt for Coder in first round of generation}]
You write custom CUDA kernels to replace the PyTorch operators in the given architecture to get speedups.
You have complete freedom to choose the set of operators you want to replace. You may decide to replace
some operators with custom CUDA kernels and leave others unchanged. You may replace multiple operators
with custom implementations, consider operator fusion opportunities (combining multiple operators into a
single kernel, for example, combining \texttt{matmul}+ \texttt{relu}), or algorithmic changes (such as online softmax).
You are only limited by your imagination.

Here is an example to show you the syntax of inline-embedding custom CUDA operators in PyTorch.

\textbf{The example given architecture is:}
\begin{lstlisting}[style=nice, language=Python]
{few_base}
\end{lstlisting}

\textbf{The example new architecture with custom CUDA kernels looks like this:}
\begin{lstlisting}[style=nice, language=Python]
{few_new}
\end{lstlisting}

\textbf{You are given the following architecture:}
\begin{lstlisting}[style=nice, language=Python]
{arch_src}
\end{lstlisting}

Optimize the architecture named \texttt{Model} with custom CUDA operators! Name your optimized output
architecture \texttt{ModelNew}. Output the new code in code blocks. Please generate real code, \emph{NOT}
pseudocode. Make sure the code compiles and is fully functional. Just output the new model code, no other
text, and \textbf{NO} testing code!
\end{tcolorbox}
\lstset{language=python}

\subsection{Prompt for Judge}
In our prompt design for the Judge agent, we place the role specification and output schema in the system prompt. Inspired by ~\citep{agashe2024agentsopenagentic,lei2025infantagentnextmultimodalgeneralistagent,chen2023agentversefacilitatingmultiagentcollaboration}, we design the prompts for correction and optimization, which are shown below.

\begin{tcolorbox}[redbox, title={Prompt for CUDA Kernel Correction}]
\ttfamily

You are a senior CUDA + PyTorch correctness auditor. Your job is to read a PyTorch reference and a CUDA candidate and report \textbf{exactly one} most critical correctness issue in the CUDA code that would cause a behavioral mismatch vs.\ the PyTorch reference. Be terse and precise.

\medskip
\textbf{Rules:}
\begin{itemize}
  \item Return \textbf{one and only one} issue --- the single highest-impact problem.
  \item Prefer semantic/correctness issues over micro-optimizations or style.
  \item If multiple issues exist, pick the one that most changes outputs or gradients.
  \item If nothing clearly wrong is found, say it explicitly.
  \item Keep each field brief; avoid extra commentary, lists, or alternatives.
\end{itemize}

\medskip
\textbf{Output format (JSON):}
\begin{tcolorbox}[colback=white, colframe=gray!50, left=2mm, right=2mm, top=1mm, bottom=1mm]
\footnotesize
\begin{verbatim}
{
  "critical_issue": "<max 20 words>",
  "why_it_matters": "<max 35 words>",
  "minimal_fix_hint": "<max 20 words>"
}
\end{verbatim}
\end{tcolorbox}

\medskip
\textbf{You are given:}

\textbf{ERROR\_LOG:}
\begin{verbatim}
{ERROR_LOG}
\end{verbatim}

\textbf{PyTorch reference (ground truth):}
\begin{verbatim}
{PYTORCH_CODE}
\end{verbatim}

\textbf{CUDA candidate (to audit):}
\begin{verbatim}
{CUDA_CODE}
\end{verbatim}

Follow the Rules and produce the JSON exactly in the specified format.

\normalfont 
\end{tcolorbox}

\begin{tcolorbox}[mybox, title={Prompt for CUDA Kernel Optimization}]
\ttfamily 

You are a senior CUDA performance engineer. Read the target GPU spec, the PyTorch
reference code, the current CUDA candidate, and the Nsight Compute metrics. Then identify \textbf{exactly one} highest-impact speed bottleneck by 3--4 most important metrics, propose \textbf{exactly one} optimisation method and propose a modification plan. Be surgical and metrics-driven.

\medskip
\textbf{Rules:}
\begin{itemize}
  \item Return \textbf{one and only one} optimisation method --- the largest expected speedup.
  \item Prefer changes that directly address measured bottlenecks (occupancy limits,
        memory coalescing, smem bank conflicts, register pressure, long/short scoreboard
        stalls, tensor-core underutilisation, etc.).
  \item Keep fields brief; avoid lists of alternatives, disclaimers, or generic advice.
\end{itemize}

\medskip
\textbf{Output format (JSON):}
\begin{tcolorbox}[colback=white, colframe=gray!50, left=2mm, right=2mm, top=1mm, bottom=1mm]
\footnotesize
\begin{verbatim}
{
  "bottleneck": "<max 30 words>",
  "optimisation method": "<max 35 words>",
  "modification plan": "<max 35 words>"
}
\end{verbatim}
\end{tcolorbox}

\medskip
\textbf{Target GPU}
\begin{verbatim}
GPU Name: {gpu_name}
Architecture: {gpu_arch}
Details:
{gpu_items}
\end{verbatim}

\textbf{PyTorch Reference}
\begin{verbatim}
{python_code}
\end{verbatim}

\textbf{CUDA Candidate}
\begin{verbatim}
{CUDA_CODE}
\end{verbatim}

\textbf{Nsight Compute metrics (verbatim)}
\begin{verbatim}
{NCU_METRICS}
\end{verbatim}

Read everything and follow the Rules exactly. Return the JSON in the specified format.

\normalfont 
\end{tcolorbox}

\subsection{Prompt for Coder from Round 2 to N}
After getting the feedback from the Judge, the Coder then corrects or optimizes the current kernel candidate based on the feedback. Prompts for correction and optimization are shown below. 

\begin{tcolorbox}[redbox, title={Prompt for Kernel Correction}]
\ttfamily 

You are a senior CUDA-extension developer.  
Your job is to \textbf{FIX} the compilation or runtime errors in the Python script shown below.

\medskip
\textbf{OUTPUT RULES (STRICT)}
\begin{verbatim}
1. Inside the block, follow exactly this order:
   1. Imports - torch, torch.nn, load_inline.
   2. source - triple-quoted CUDA string(s) (kernel + host wrapper).
   3. cpp_src - prototypes for all kernels you expose.
   4. One load_inline call per kernel group.
   5. class ModelNew(nn.Module) - mirrors original inputs/outputs but calls
      your CUDA kernels.
2. Do NOT include testing code, if __name__ == "__main__", or extra prose.
\end{verbatim}

\medskip
\textbf{ERROR LOG}
\begin{tcolorbox}[colback=white, colframe=gray!50, left=2mm, right=2mm, top=1mm, bottom=1mm]
\footnotesize
\begin{verbatim}
{ERROR_LOG}
\end{verbatim}
\end{tcolorbox}

\textbf{OLD CODE (read-only)}
\begin{tcolorbox}[colback=white, colframe=gray!50, left=2mm, right=2mm, top=1mm, bottom=1mm]
\footnotesize
\begin{verbatim}
{CUDA_CODE}
\end{verbatim}
\end{tcolorbox}

\textbf{Main Critical Problem}
\begin{tcolorbox}[colback=white, colframe=gray!50, left=2mm, right=2mm, top=1mm, bottom=1mm]
\footnotesize
\begin{verbatim}
{Problem}
\end{verbatim}
\end{tcolorbox}

\medskip
\textbf{Output Section (to be generated):}
\begin{tcolorbox}[colback=white, colframe=gray!50, left=2mm, right=2mm, top=1mm, bottom=1mm]
\footnotesize
\begin{verbatim}
# <your corrected code>
\end{verbatim}
\end{tcolorbox}

\normalfont 
\end{tcolorbox}

\begin{tcolorbox}[mybox, title={Prompt for Kernel Optimization}]
\ttfamily 

\textbf{Target GPU}
\begin{verbatim}
GPU Name: {gpu_name}
Architecture: {gpu_arch}
Details:
{gpu_items}
\end{verbatim}

You are a CUDA-kernel optimization specialist.

Analyze the provided architecture and \textbf{strictly apply the following STRATEGY} to produce an improved CUDA kernel.

\begin{tcolorbox}[colback=white, colframe=gray!50, left=2mm, right=2mm, top=1mm, bottom=1mm]
\footnotesize
\begin{verbatim}
{CUDA_CODE}
\end{verbatim}
\end{tcolorbox}

\textbf{Optimization instructions:}
\begin{verbatim}
{optimization_suggestion}
\end{verbatim}

\medskip
\textbf{GOAL}
\begin{verbatim}
----
- Improve latency and throughput on the target GPU.
- Maintain correctness within atol=1e-4 or rtol=1e-4.
- Preserve the public Python API (same inputs/outputs, shapes, dtypes).
\end{verbatim}

\medskip
\textbf{OUTPUT RULES (STRICT)}
\begin{verbatim}
1. Inside the block, follow exactly this order:
   1. Imports - torch, torch.nn, load_inline.
   2. source - triple-quoted CUDA string(s) (kernel + host wrapper).
   3. cpp_src - prototypes for all kernels you expose.
   4. One load_inline call per kernel group.
   5. class ModelNew(nn.Module) - mirrors original inputs/outputs but calls
      your CUDA kernels.
2. Do NOT include testing code, if __name__ == "__main__", or extra prose.
\end{verbatim}

\medskip
\textbf{Output Section (to be generated):}
\begin{tcolorbox}[colback=white, colframe=gray!50, left=2mm, right=2mm, top=1mm, bottom=1mm]
\footnotesize
\begin{verbatim}
# <your corrected code>
\end{verbatim}
\end{tcolorbox}

\normalfont 
\end{tcolorbox}

\section{Detail for the NCU metrics}

\subsection{Why choose NCU subset metrics?}\label{app:why_subset}
We find that exposing large models to the full NCU metric set overwhelms them, reducing the accuracy and stability of their optimization suggestions and degrading Judge output quality. We illustrate this with following specific case study.
\begin{figure}[H]
  \centering
  \includegraphics[width=0.8\linewidth]{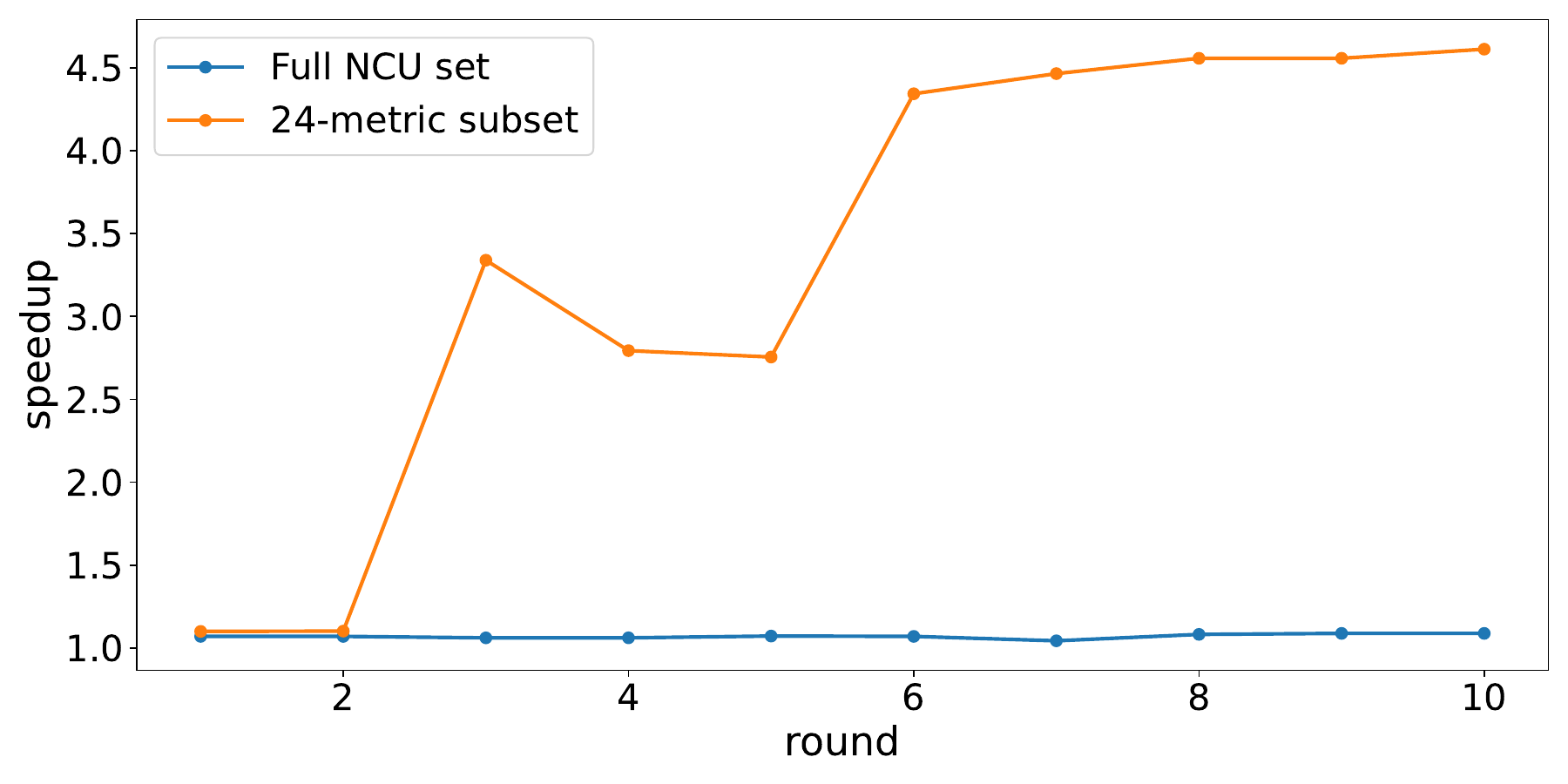}
  \caption{Generated kernel performance comparison between Full-metrics input(blue line) vs. Metrics subset input(orange line) for Judge agent on Level 2 task 51. }
  \label{fig:why_subset}
\end{figure}

As shown in Figure~\ref{fig:why_subset}, we compare two configurations on a representative task. (i) The Judge agent receives {\it all} raw NCU metrics in each round and provides feedback to the Coder agent. (ii) The Judge agent receives only the 24 key metrics selected by our method and captures only 3-4 critical metrics in each round to generate optimization suggestion.

Within 10 iterative rounds, the Judge agent using full NCU metrics fails to guide the Coder agent toward better kernels. The performance of the generated kernels remains almost unchanged, despite repeated interactions. In contrast, when the Judge agent is restricted to use subset metrics, the performance of generated kernels varies across rounds and is often improved. This process eventually leads to a substantial speedup.

In addition, we carefully reviewed how the Judge identified bottlenecks and proposed optimizations in both configurations, full set vs. 24-subset. We select one Judge's response from the optimization round to better illustrate the difference in judge output, with full NCU and with a subset of NCU as judge input.

\begin{tcolorbox}[mybox, title={Response by using the full NCU metrics set}]
\ttfamily 
\begin{verbatim}
{"bottleneck": "Long-scoreboard stalls from re-loading GEMM output out of DRAM
(gemm_out round-trip dominates cycles).",

"optimisation method": "Fuse the Linear GEMM and the current epilogue into one FP16
tensor-core CUTLASS kernel that performs subtract, row-mean, GELU and residual add
in-register, avoiding global-memory re-reads.",

"modification plan": "Write CUTLASS custom epilogue functor; use cublasLt or cutlass::
gemm_universal to compute GEMM+A, keep accumulators in registers, warp-reduce to mean,
apply GELU, add x_orig, directly store final output; expose via new PyTorch extension."}
\end{verbatim}
\normalfont
\end{tcolorbox}

\begin{tcolorbox}[redbox, title={Response by using the 24 key NCU subset}]
\ttfamily 
\begin{verbatim}
{"bottleneck": "Kernel is DRAM-bound (102.9 % peak); 57 % long-scoreboard stalls from
second pass reading original_x.",

"optimisation method": "Fuse the second loop: cache each thread’s original_x elements
in registers during diff accumulation, then add gelu_val and store, eliminating the 
extra B×F global read.",

"modification plan": "Replace two for-loops with one: load linear_out, subtract, and
original_x together, keep original_x[16] in per-thread array, finish reduction, add 
gelu_val, write to output; ~4 MB traffic per batch, >30 % speedup."}
\end{verbatim}
\normalfont
\end{tcolorbox}
Based on these two responses,we find that judge with full set NCU mertics tends to misidentify the true bottleneck.The judge with full set NCU metrics attributes the bottleneck to re-loading gemm\_out and recommends a monolithic CUTLASS epilogue that performs row-mean/GELU/residual in registers. This diagnosis is misaligned with our kernel’s access pattern and is hard to realize for general shapes due to cross-tile aggregation. In contrast, the judge with 24-key subset correctly identifies a DRAM-bound kernel dominated by the second pass over x\_orig, and proposes a one-pass rewrite that caches x\_orig in registers during the first traversal and writes back after GELU, eliminating an entire B×F global memory read. This change is lightweight, architecture-agnostic, and yields consistent speedups (e.g., about 4 MB less traffic per batch, more than 30\% in our setting).

\subsection{Top-20 NCU metrics Example}\label{app:top-20}

 This section reports, in several example tasks, the Top-20 Nsight Compute (NCU) metrics most correlated with runtime, ranked by the absolute value of the Pearson correlation coefficient. Here, runtime refers to the kernel’s execution time. When the correlation coefficient is positive, larger metric values typically imply longer execution time; when it is negative, larger metric values typically imply shorter execution time. All metric names follow their original name in NCU.

\scriptsize
\begin{longtable}{>{\raggedright\arraybackslash}p{0.63\textwidth}
                  S[table-format=+1.6]
                  S[table-format=1.6]}
\caption{Task-Conv2D: Pearson correlation with runtime (Top-20).}
\label{tab:app-conv2d-top20}\\
\toprule
\textbf{Metric Name} & {\textbf{Correlation}} & {\textbf{Abs Correlation}} \\
\midrule
\endfirsthead
\toprule
\textbf{Metric Name} & {\textbf{Correlation}} & {\textbf{Abs Correlation}} \\
\midrule
\endhead
\midrule
\multicolumn{3}{r}{\small Continued on next page} \\
\endfoot
\bottomrule
\endlastfoot

\Metric{sm\_\_cycles\_active.avg}                                  &  1.000000 & 1.000000 \\
\Metric{gpc\_\_cycles\_elapsed.max}                                &  1.000000 & 1.000000 \\
\Metric{launch\_\_occupancy\_limit\_shared\_mem}                    &  0.945507 & 0.945507 \\
\Metric{dram\_\_bytes.sum.per\_second}                              & -0.924251 & 0.924251 \\
\Metric{gpu\_\_dram\_throughput.avg.pct\_of\_peak\_sustained\_elapsed} & -0.924155 & 0.924155 \\
\Metric{smsp\_\_inst\_executed.avg}                                 &  0.916287 & 0.916287 \\
\Metric{smsp\_\_inst\_executed.sum}                                 &  0.916287 & 0.916287 \\
\Metric{smsp\_\_inst\_issued.avg}                                   &  0.916262 & 0.916262 \\
\Metric{smsp\_\_inst\_issued.sum}                                   &  0.916262 & 0.916262 \\
\Metric{lts\_\_t\_sector\_hit\_rate.pct}                            &  0.839237 & 0.839237 \\
\Metric{smsp\_\_sass\_average\_branch\_targets\_threads\_uniform.pct} &  0.810334 & 0.810334 \\
\Metric{lts\_\_throughput.avg.pct\_of\_peak\_sustained\_elapsed}    & -0.787261 & 0.787261 \\
\Metric{smsp\_\_inst\_executed\_op\_branch.sum}                     &  0.746483 & 0.746483 \\
\Metric{launch\_\_grid\_size}                                       &  0.745917 & 0.745917 \\
\Metric{l1tex\_\_t\_sector\_hit\_rate.pct}                          &  0.728356 & 0.728356 \\
\Metric{gpc\_\_cycles\_elapsed.avg.per\_second}                     &  0.728053 & 0.728053 \\
\Metric{dram\_\_cycles\_elapsed.avg.per\_second}                    &  0.665784 & 0.665784 \\
\Metric{launch\_\_waves\_per\_multiprocessor}                       &  0.627478 & 0.627478 \\
\Metric{launch\_\_thread\_count}                                    &  0.627478 & 0.627478 \\
\Metric{launch\_\_shared\_mem\_per\_block\_static}                  & -0.610501 & 0.610501 \\

\end{longtable}
\scriptsize
\begin{longtable}{>{\raggedright\arraybackslash}p{0.63\textwidth}
                  S[table-format=+1.6]
                  S[table-format=1.6]}
\caption{Task-SpMM: Pearson correlation with runtime (Top-20).}
\label{tab:app-spmm-top20}\\
\toprule
\textbf{Metric Name} & {\textbf{Correlation}} & {\textbf{Abs Correlation}} \\
\midrule
\endfirsthead
\toprule
\textbf{Metric Name} & {\textbf{Correlation}} & {\textbf{Abs Correlation}} \\
\midrule
\endhead
\midrule
\multicolumn{3}{r}{\small Continued on next page} \\
\endfoot
\bottomrule
\endlastfoot

\Metric{gpc\_\_cycles\_elapsed.max}                                &  0.999993 & 0.999993 \\
\Metric{sm\_\_cycles\_active.avg}                                  &  0.998432 & 0.998432 \\
\Metric{gpu\_\_compute\_memory\_request\_throughput.avg.pct\_...}  & -0.967284 & 0.967284 \\
\Metric{gpu\_\_compute\_memory\_throughput.avg.pct\_of\_peak\_...} & -0.964455 & 0.964455 \\
\Metric{lts\_\_t\_sector\_hit\_rate.pct}                           &  0.951201 & 0.951201 \\
\Metric{dram\_\_bytes.sum.per\_second}                              & -0.926134 & 0.926134 \\
\Metric{gpu\_\_dram\_throughput.avg.pct\_of\_peak\_sustained\_...} & -0.925856 & 0.925856 \\
\Metric{l1tex\_\_throughput.avg.pct\_of\_peak\_sustained\_active}  &  0.871262 & 0.871262 \\
\Metric{sm\_\_inst\_executed.avg.per\_cycle\_elapsed}              & -0.837675 & 0.837675 \\
\Metric{smsp\_\_issue\_inst0.avg.pct\_of\_peak\_sustained\_active} &  0.837284 & 0.837284 \\
\Metric{smsp\_\_issue\_active.avg.pct\_of\_peak\_sustained\_...}   & -0.837284 & 0.837284 \\
\Metric{smsp\_\_issue\_active.avg.per\_cycle\_active}              & -0.837283 & 0.837283 \\
\Metric{sm\_\_inst\_issued.avg.per\_cycle\_active}                 & -0.836185 & 0.836185 \\
\Metric{sm\_\_inst\_issued.avg.pct\_of\_peak\_sustained\_active}   & -0.836185 & 0.836185 \\
\Metric{sm\_\_inst\_executed.avg.per\_cycle\_active}               & -0.836160 & 0.836160 \\
\Metric{sm\_\_instruction\_throughput.avg.pct\_of\_peak\_sust...}  & -0.806478 & 0.806478 \\
\Metric{smsp\_\_average\_warp\_latency\_per\_inst\_issued.ratio}   &  0.802793 & 0.802793 \\
\Metric{smsp\_\_average\_warps\_active\_per\_inst\_executed.ratio} &  0.802777 & 0.802777 \\
\Metric{derived\_\_smsp\_\_inst\_executed\_op\_branch\_pct}        & -0.728768 & 0.728768 \\
\Metric{smsp\_\_warps\_eligible.avg.per\_cycle\_active}            & -0.630772 & 0.630772 \\
\end{longtable}
\normalsize
\subsection{Key Subset of 24 NCU Metrics}\label{app:key24}

The table below lists the exact 24 metrics in our NCU key subset, as a result of Algorithm~\ref{alg::alg1},~\ref{alg::alg2}

\scriptsize
\begin{longtable}{rl}
\caption{The 24-metric key subset.}
\label{tab:key24}\\
\toprule
\textbf{\#} & \textbf{Metric Name} \\
\midrule
\endfirsthead
\toprule
\textbf{\#} & \textbf{Metric Name} \\
\midrule
\endhead
\midrule
\multicolumn{2}{r}{\small Continued on next page}\\
\endfoot
\bottomrule
\endlastfoot

1  & \Metric{sm\_\_cycles\_active.avg} \\
2  & \Metric{sm\_\_warps\_active.avg.pct\_of\_peak\_sustained\_active} \\
3  & \Metric{launch\_\_occupancy\_limit\_blocks} \\
4  & \Metric{launch\_\_occupancy\_limit\_registers} \\
5  & \Metric{launch\_\_occupancy\_limit\_shared\_mem} \\
6  & \Metric{launch\_\_registers\_per\_thread} \\
7  & \Metric{sm\_\_inst\_executed.sum} \\
8  & \Metric{sm\_\_inst\_executed\_pipe\_fp32.avg.pct\_of\_peak\_sustained\_active} \\
9  & \Metric{sm\_\_inst\_executed\_pipe\_tensor.avg.pct\_of\_peak\_sustained\_active} \\
10 & \Metric{dram\_\_bytes\_read.sum} \\
11 & \Metric{dram\_\_bytes\_write.sum} \\
12 & \Metric{dram\_\_throughput.avg.pct\_of\_peak\_sustained\_elapsed} \\
13 & \Metric{dram\_\_bytes.sum.per\_second} \\
14 & \Metric{gpu\_\_dram\_throughput.avg.pct\_of\_peak\_sustained\_elapsed} \\
15 & \Metric{l1tex\_\_t\_sector\_hit\_rate.pct} \\
16 & \Metric{l1tex\_\_throughput.avg.pct\_of\_peak\_sustained\_active} \\
17 & \Metric{lts\_\_t\_sector\_hit\_rate.pct} \\
18 & \Metric{lts\_\_throughput.avg.pct\_of\_peak\_sustained\_active} \\
19 & \Metric{smsp\_\_warp\_issue\_stalled\_memory\_dependency\_per\_warp\_active.pct} \\
20 & \Metric{smsp\_\_warp\_issue\_stalled\_short\_scoreboard\_per\_warp\_active.pct} \\
21 & \Metric{smsp\_\_warp\_issue\_stalled\_long\_scoreboard\_per\_warp\_active.pct} \\
22 & \Metric{smsp\_\_warp\_issue\_stalled\_barrier\_per\_warp\_active.pct} \\
23 & \Metric{smsp\_\_warp\_issue\_stalled\_branch\_resolving\_per\_warp\_active.pct} \\
24 & \Metric{smsp\_\_sass\_average\_branch\_targets\_threads\_uniform.pct} \\
\end{longtable}
\normalsize
\section{CUDA-L1}\label{app:cuda-l1}
In our replication efforts, we found that the authors of CUDA-L1 released only the final, generated kernels for each task. After carefully studying these cases, we identified several interesting findings.

First, We found that CUDA-L1 tends to emphasize PyTorch-level optimizations rather than generating and refining custom CUDA kernels. This pattern also emerged as the most frequent issue in their provided case. Although CUDA-L1 reports the top-10 cases with the largest speedups, our review shows that nine of these ten final solutions do not use custom CUDA kernels; instead, they rely heavily on official PyTorch implementations. We show several cases of their results below.

This is the top-ranked entry in their \emph{KernelBench Tasks Ranked by RL-CUDA1 Acceleration (Top-10)}: Level 2 Task 83, with a reported $120.3\times$ speedup. It contains no CUDA kernel.
\begin{tcolorbox}[mybox, title={Level 2 Task 83 — Reported $120.3\times$ Speedup}]
\ttfamily
\begin{verbatim}
import torch
import torch.nn as nn

class ModelNew(nn.Module):
    """
    Optimized implementation of a model that performs a 3D convolution,
    applies Group Normalization, minimum, clamp, and dropout.
    
    Args:
        in_channels (int): Number of input channels
        out_channels (int): Number of output channels
        kernel_size (int or tuple): Size of the convolving kernel
        groups (int): Number of groups for GroupNorm
        min_value (float): Minimum value for clamp operation
        max_value (float): Maximum value for clamp operation
        dropout_p (float): Dropout probability
    """
    def __init__(self, in_channels, out_channels, kernel_size, groups, min_value, 
    max_value, dropout_p):
        super(ModelNew, self).__init__()
        # Store the original layers for parameter compatibility
        self.conv = nn.Conv3d(in_channels, out_channels, kernel_size)
        self.norm = nn.GroupNorm(groups, out_channels)
        self.dropout = nn.Dropout(dropout_p)
        self.min_value = min_value
        self.max_value = max_value
        self.dropout_p = dropout_p
        
        # Pre-compute convolution parameters for output shape calculation
        if isinstance(kernel_size, int):
            self.kernel_size = (kernel_size, kernel_size, kernel_size)
        else:
            self.kernel_size = kernel_size
            
        self.stride = self.conv.stride
        self.padding = self.conv.padding
        self.dilation = self.conv.dilation

    def forward(self, x):
        x = self.conv(x)
        x = self.norm(x)
        x = torch.minimum(x, torch.tensor(self.min_value, device=x.device))
        x = torch.clamp(x, min=self.min_value, max=self.max_value)
        x = self.dropout(x)
        return x

# Keep ALL hyperparameters EXACTLY as shown in the reference implementation
batch_size = 128
in_channels = 3
out_channels = 16
depth, height, width = 16, 32, 32
kernel_size = 3
groups = 8
min_value = 0.0
max_value = 1.0
dropout_p = 0.2

def get_inputs():
    return [torch.randn(batch_size, in_channels, depth, height, width)]

def get_init_inputs():
    return [in_channels, out_channels, kernel_size, groups, min_value, max_value, 
    dropout_p]
\end{verbatim}
\normalfont
\end{tcolorbox}

The second-ranked case is Level-1 Task 12 (Matmul with diagonal matrices), with a reported 64.4× speedup, which also contains no CUDA kernel:
\begin{tcolorbox}[redbox, title={Level 1 Task 12 — Reported $64.4\times$ Speedup}]
\ttfamily
\begin{verbatim}
# diag_mm_compare.py
import time
import math
import torch
import torch.nn as nn
import torch.nn.functional as F

# -------------------------------
# Reference implementation
# -------------------------------
class Model(nn.Module):
    """
    Simple model that performs a matrix multiplication of a diagonal matrix with another
    matrix.
    C = diag(A) * B
    """
    def __init__(self):
        super(Model, self).__init__()

    def forward(self, A, B):
        """
        Args:
            A (torch.Tensor): 1D tensor, diagonal entries. Shape: (N,)
            B (torch.Tensor): 2D tensor. Shape: (N, M)
        Returns:
            torch.Tensor: (N, M)
        """
        return torch.diag(A) @ B


# -------------------------------
# Optimized implementation
# -------------------------------
class ModelNew(nn.Module):
    """
    Optimized model that performs a matrix multiplication of a diagonal matrix with another
    matrix.
    C = diag(A) * B
    """
    def __init__(self):
        super(ModelNew, self).__init__()

    def forward(self, A, B):
        """
        Args:
            A (torch.Tensor): 1D tensor, diagonal entries. Shape: (N,)
            B (torch.Tensor): 2D tensor. Shape: (N, M)
        Returns:
            torch.Tensor: (N, M)
        """
        # Equivalent to torch.diag(A) @ B, but avoids forming the full diagonal matrix
        return B * A.unsqueeze(1)


# -------------------------------
# Hyperparameters & inputs
# -------------------------------
M = 4096
N = 4096

def get_inputs(device=None, dtype=torch.float32):
    A = torch.randn(N, device=device, dtype=dtype)
    B = torch.randn(N, M, device=device, dtype=dtype)
    return [A, B]

def get_init_inputs():
    return []  # No special initialization inputs needed
\end{verbatim}
\normalfont
\end{tcolorbox}
In addition, we observed many reported speedups that are effectively equal to one (clustered around 1.00, typically within ±5\%). A closer inspection shows that, in these cases, the system falls back to the original PyTorch operator when the custom kernel fails to compile, which naturally yields no measurable speedup.

For example, below is the forward method from the final solution for KernelBench Level-1 Task 3 generated by CUDA-L1. This code get from the CUDA-L1's official Github. We observe that the method first attempts to call a \emph{custom CUDA kernel}; however, upon any compilation failure or exception, it immediately falls back to \texttt{torch.bmm(A, B)}. Crucially, \texttt{torch.bmm(A, B)} is exactly the operator that this task asks to be replaced by a custom kernel, meaning the fallback undermines the task's objective. This explains why the reported speedup is only 1.006×.
\begin{tcolorbox}[mybox, title={Level 1 Task 3 — Reported $1.006\times$ Speedup}]
\ttfamily
\begin{verbatim}
    def forward(self, A: torch.Tensor, B: torch.Tensor) -> torch.Tensor:
        """
        Performs batched matrix multiplication.

        Args:
            A: Input tensor of shape (batch_size, m, k).
            B: Input tensor of shape (batch_size, k, n).

        Returns:
            C: Output tensor of shape (batch_size, m, n).
        """
        # Fall back to torch.bmm if CUDA module failed to load
        if ModelNew._cuda_module is None:
            return torch.bmm(A, B)
        
        # Check if inputs are on CUDA
        if not A.is_cuda or not B.is_cuda:
            A = A.cuda() if not A.is_cuda else A
            B = B.cuda() if not B.is_cuda else B
        
        # Ensure inputs are contiguous and float32
        A = A.contiguous().float()
        B = B.contiguous().float()
        
        # Use custom CUDA kernel
        try:
            result = ModelNew._cuda_module.batched_matmul(A, B)
            if not A.is_cuda:
                result = result.cpu()
            return result
        except Exception as e:
            print(f"Error in custom kernel: {e}, falling back to torch.bmm")
            return torch.bmm(A, B)
\end{verbatim}
\normalfont
\end{tcolorbox}

\section{Details of Benchmark}
\subsection{KernelBench}\label{KernelBench}
\textbf{KernelBench} is a standardized benchmark designed to evaluate the capability of large language models (LLMs) in CUDA kernel generation and optimization.
It consists of 270 tasks across four levels of increasing difficulty, of which Levels 1–3 (250 tasks in total) are commonly adopted for evaluation. Each task provides a PyTorch reference implementation together with fixed input–output specifications, enabling automated correctness and performance validation.

\begin{itemize}
\item \textbf{Level 1 (Basic Operators):} Contains simple, low-level operators such as matrix multiplication, element-wise operations, and reductions. These tasks primarily test the ability to generate functionally correct CUDA kernels.
\item \textbf{Level 2 (Composite Operations):} Involves multi-step operator combinations, requiring the model to compose multiple CUDA primitives and manage intermediate memory efficiently. These tasks test the capacity for more complex code synthesis.
\item \textbf{Level 3 (End-to-End Models):} Includes challenging kernels derived from full neural network architectures such as AlexNet, VGG, and ResNet components. These tasks assess the ability to produce efficient, large-scale kernels under realistic deep learning workloads.
\item \textbf{Level 4 (Optional):} The full benchmark also defines an advanced level with additional research-oriented tasks, but this is less frequently adopted due to its complexity and lack of standardized evaluation setups.
\end{itemize}

KernelBench has become a widely used benchmark in recent work on LLM-based code generation \citep{deepreinforce2025cudal1, kevin-multi-turn-rl, lange2025robustagenticcudakernel}, as it provides a controlled and reproducible environment to measure both \emph{correctness} (functional equivalence to PyTorch) and \emph{efficiency} (execution speed relative to PyTorch).
In our study, we adopt all Level 1–3 tasks, following prior work, to ensure fair comparison across baselines.

\subsection{Our stratified random subset \(\mathcal{D}^{*}\)}
\label{subset}
While our main evaluation is conducted on the full KernelBench Level 1–3 benchmark 
(250 tasks in total), we additionally construct a stratified subset $\mathcal{D}^{*}$ 
to enable detailed analysis and fair comparison, while reducing computation cost in experiments.  

The construction of $\mathcal{D}^{*}$ follows two principles:  
(1) \textbf{Coverage across difficulty levels.} Since KernelBench is stratified by 
increasing task complexity (Level~1: single-operator tasks, Level~2: multi-step fused operators, 
Level~3: full network components), we ensure that the sampled subset preserves the relative 
distribution of difficulty.  
(2) \textbf{Diversity of task types.} Within each level, we sample tasks uniformly across different 
operator categories (e.g., elementwise ops, reductions, convolutions, fused blocks) so that the 
subset remains representative of the overall benchmark.  

Concretely, we perform stratified random sampling with a fixed 10\% ratio for each level, 
resulting in a subset of 10 tasks from Level~1, 10 tasks from Level~2, and 5 tasks from Level~3, 
for a total of 25 tasks. 
For reproducibility, the exact task IDs included in $\mathcal{D}^{*}$ are:  
\begin{itemize}
    \item \textbf{Level 1 (10 tasks):} 13, 10, 16, 29, 35, 72, 7, 89, 93, 34  
    \item \textbf{Level 2 (10 tasks):} 17, 19, 40, 3, 13, 21, 38, 28, 26, 34  
    \item \textbf{Level 3 (5 tasks):} 5, 18, 32, 41, 21
\end{itemize}

\end{document}